\documentclass[11pt]{article}
\usepackage{coling}
\usepackage{times}
\usepackage{latexsym}
\usepackage{array}
\usepackage{spverbatim}
\usepackage{soul,color}
\usepackage{colortbl}
\usepackage{booktabs}
\usepackage{multirow}
\usepackage{float}
\usepackage[T1]{fontenc}
\usepackage{graphicx} 
\usepackage[utf8]{inputenc}
\usepackage{comment}
\usepackage{microtype}
\usepackage{amsmath}
\usepackage{enumitem}
\usepackage{inconsolata}
\usepackage{soul,color}
\usepackage{subcaption}
\usepackage{balance}

\title{Leveraging Language Models for Summarizing Mental State Examinations: A Comprehensive Evaluation and Dataset Release}

\author{
  \textbf{Nilesh Kumar Sahu\textsuperscript{1}\textsuperscript{$*$}\textsuperscript{$\Delta$}},
  \textbf{Manjeet Yadav\textsuperscript{1}\textsuperscript{$*$}},
  \textbf{Mudita Chaturvedi\textsuperscript{2}},
  \textbf{Snehil Gupta\textsuperscript{2}},
  \textbf{Haroon R Lone\textsuperscript{1}}
\\
\\
  \textsuperscript{1}IISER Bhopal India,
  \textsuperscript{2}AIIMS Bhopal India
\\
}

\begin{document}
\maketitle
\def\thefootnote{*}\footnotetext{Equal contributions.}
\def\thefootnote{$\Delta$}\footnotetext{Corresponding author.}

\makeatletter
\renewcommand{\thefootnote}{\@arabic\c@footnote}
\makeatother

\begin{abstract}
Mental health disorders affect a significant portion of the global population, with diagnoses primarily conducted through Mental State Examinations (MSEs). MSEs serve as structured assessments to evaluate behavioral and cognitive functioning across various domains, aiding mental health professionals in diagnosis and treatment monitoring. However, in developing countries, access to mental health support is limited, leading to an overwhelming demand for mental health professionals. Resident doctors often conduct initial patient assessments and create summaries for senior doctors, but their availability is constrained, resulting in extended patient wait times.

This study addresses the challenge of generating concise summaries from MSEs through the evaluation of various language models.
Given the scarcity of relevant mental health conversation datasets, we developed a 12-item descriptive MSE questionnaire and collected responses from 405 participants, resulting in 9720 utterances covering diverse mental health aspects. Subsequently, we assessed the performance of five well-known pre-trained summarization models, both with and without fine-tuning, for summarizing MSEs. Our comprehensive evaluation, leveraging metrics such as ROUGE, SummaC, and human evaluation, demonstrates that language models can generate automated coherent MSE summaries for doctors. With this paper, we release our collected conversational dataset and trained models publicly for the mental health research community.

\end{abstract}

\maketitle

\section{Introduction}
Mental health disorders are prevalent worldwide. A recent study shows that one in every eight people suffers from some mental health disorder~\citep{world2022world}. Usually, mental health disorders are diagnosed in clinical settings with Mental State Examination (MSE).
An MSE is a structured assessment of the behavioral and cognitive functioning of an individual suffering from a mental health disorder \citep{martin1990mental,voss2019mental}. It aids in comprehending psychological functioning across multiple domains, including mood, thoughts, perception, cognition, etc. Mental health professionals (i.e., psychiatrists and psychologists) utilize MSEs at different treatment stages (prior, during, or after) to grasp the onset of mental health disorders, assess the effectiveness of therapy sessions, and evaluate the progress of treatment.

In developing countries, mental health support is limited, with only a few mental health professionals available for a large number of patients \citep{majumdar2022covid, rojas2019improving,saraceno2007barriers}. Resident (junior) doctors, supervised by senior doctors, are commonly employed to manage the demand. The primary responsibility of such junior doctors is to conduct initial patient assessments through structured MSEs and create concise summaries of issues and symptoms for senior doctors. Reviewing these summaries reduces evaluation time for senior doctors, allowing them more time to focus on treatment planning. 



Developing an automated system for initial assessment and summary generation would be pivotal in simulating an Artificial intelligence (AI)-driven junior doctor. The system would conduct MSEs and generate concise summaries of the MSE for the attending senior doctor~\citep{jain2022survey}. 
The automated system will consist of two main parts: (i) a user interface for gathering user responses to MSE questions and (ii) an AI module for summarizing those responses. This study focuses on the latter by evaluating various language models to determine their effectiveness in generating concise summaries from MSEs. Summarizing accurately and concisely using pre-trained language models is challenging due to a lack of relevant mental health conversation datasets \citep{qiu2023smile} and the significant shift in content from non-mental to mental health topics. To tackle these challenges, we first developed a 12-item descriptive MSE and collected data by conducting MSEs with 405 participants. Collecting responses on a 12-item questionnaire was the most challenging step in our study, as it took around 20-25 minutes to respond to the questions for each participant. The unique design of the questionnaire, capturing diverse aspects such as mood, social life, family dynamics, etc, makes the collected dataset valuable for the research community to answer a range of mental health research questions.  
Next, using our dataset, we assessed the performance of five well-known pre-trained language models with and without fine-tuning for summarizing MSEs. The selected language models are known for their state-of-the-art performance for text summarization. 
Our comprehensive evaluation, based on metrics such as ROUGE scores, SummaC score, and human evaluation, indicates that fine-tuning pre-trained language models, even with limited training data, improves the generation of accurate and coherent summaries. Notably, the best fine-tuned models outperform existing baseline language models, achieving ROUGE-1 and ROUGE-L scores of 0.829 and 0.790, respectively. With this paper, we release our collected conversational dataset\footnote{\url{https://huggingface.co/SIR-Lab/MSE_Summarizer}} and trained models publicly for the mental health research community

\section{Related Works}
\subsection{Dialogue summarization}

Models like BART~\citep{lewis2020bart} \& GPT-3~\citep{radford2018improving}, with their numerous parameters, demonstrate exceptional performance across various general-purpose tasks. However, their training primarily relies on knowledge-based resources such as books, web documents, and academic papers. Nonetheless, they often require additional domain-specific conversation data to understand dialogues better. The lack of publicly available appropriate data sets creates a challenge for generating abstractive summaries. To overcome this challenge, Samsung research team \citep{gliwa2019samsum} made their dataset publicly available. Furthermore, ~\citep{zhong2022dialoglm} introduced a pre-training framework for understanding and summarizing long dialogues. Recently introduced PEGASUS~\citep{zhang2020pegasus}, an innovative summarization framework founded upon a transformer-based encoder-decoder architecture, represents the latest frontier in this evolving landscape. Similarly, \citep{yun2023fine} enhanced routine functions for customer service representatives by employing a fine-tuning method for dialogue summarization. However, medical dialogues present unique challenges due to the inclusion of critical information such as medical history, the context of the doctor, and the severity of patient responses, necessitating specialized approaches beyond those employed in typical dialogue processing.

\subsection{Medical dialogue summarization}
Recent advancements in automatic medical dialogue summarization have propelled the field forward significantly. Notably, both LSTM and transformer models have demonstrated the capability to generate concise summaries from doctor-patient conversations~\citep{krishna2021generating,srivastava2022counseling,song2024clinically}. For example, \citep{song2024clinically}  generated summaries from social media timeline and \citep{srivastava2022counseling} generated summaries from counseling sessions. 
Furthermore, pre-trained transformer models have been leveraged to summarize such conversations from transcripts directly~\citep{zhang2021leveraging,michalopoulos2022medicalsum,enarvi2020generating}.

In addition, the hierarchical encoder-tagger model has emerged as a promising approach, producing summaries by identifying \& extracting meaningful utterances, mainly focusing on problem statements and treatment recommendations~\citep{song2020summarizing}. However, it is important to note that these models are typically trained on brief, general physician-patient conversations. In contrast, conversations in the psychological domain tend to be longer, with more detailed patient responses. Understanding the nuances of behavior \& thinking patterns becomes crucial for accurate diagnosis in such contexts. \citep{yao2022d4} addressed this challenge by fine-tuning a pre-trained language model to generate symptom summaries from psychiatrist-patient conversations on a Chinese dataset. 

To enhance the applicability of language models in the mental health domain, \citep{yang2023mentalllama} curated an extensive mental health dataset from social media to train MentaLLaMA. Similarly, \citep{ji2021mentalbert} utilized various datasets focused on depression, anxiety, and suicidal ideation from diverse social media platforms to train models like MentalBERT and MentalRoBERTa. However, it is worth noting that fine-tuning or deploying such models on low-computational machines poses challenges. Techniques such as model pruning or quantization can be employed to reduce the model size. However, these methods may introduce compatibility issues with hardware accelerators or deployment platforms~\citep{kuzmin2024pruning,dery2024everybody}. Additionally, they may compromise the model's efficiency, potentially impacting its performance.

Several benchmarks have been established to assess the quality of generated summaries based on various criteria~\citep{joseph2024factpico, cai2023medbench}. However, current summarization models producing factually inconsistent summaries are unsuitable for real-world applications~\citep{zablotskaia2023calibrating,chen2023evaluating}. Hallucination, in particular, is a significant issue with current models~\citep{zablotskaia2023calibrating}. Although efforts have been made to improve consistency, such as those by \citep{zablotskaia2023calibrating}, these approaches cannot completely guarantee the absence of hallucination.
Therefore, achieving a balance between quality, simplicity, and factuality in generated summaries remains a challenge~\citep{joseph2024factpico, dixit2023improving,feng2023improving}.

\begin{figure}
    \centering
    \includegraphics[width=1\linewidth]{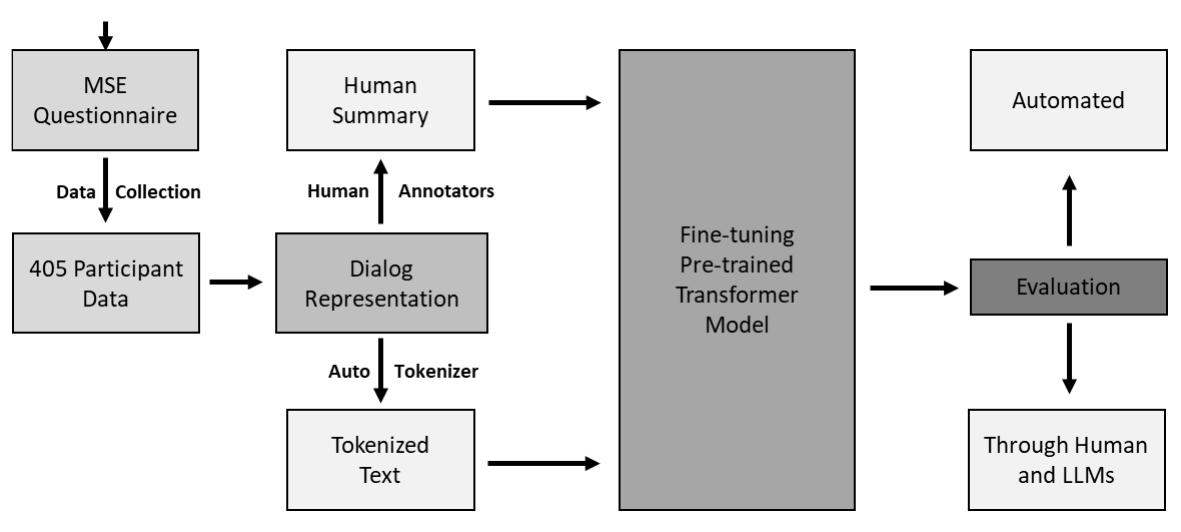}
    \caption{Methodology flowchart}
    \label{fig:methodology}
\end{figure}

\section{Methodology}
Figure~\ref{fig:methodology} provides a high-level overview of the methodology. Following is a detailed description of the methodology sub-components. 

\subsection{MSE questionnaire design}
Due to the absence of a standardized MSE questionnaire, we created a preliminary version tailored to students, encompassing key components like \textbf{socialness}, \textbf{mood}, \textbf{attention}, \textbf{memory}, \textbf{frustration tolerance}, and \textbf{social support} after several meetings with student counselors, psychologists, and going through publicly available counseling videos on YouTube. This process yielded an 18-item questionnaire. Subsequently, we sought the expertise of clinical psychiatrists to refine the questionnaire further. Their valuable insights were instrumental in vetting the relevance, resulting in a finalized version of the MSE comprising 12 questions. Finally, the questionnaire was validated by a separate team of four psychiatrists based on item accuracy, language clarity, and reliability, following the guidelines outlined  by Jones et al. \citep{jones1995consensus} and Gupta et al. \citep{gupta2022development}. Tables {\bf \ref{appendix: mse_questionnaire validation}} and  \textbf{\ref{tab:MSE Questionnaire}} in the appendix lists the questionnaire validation scores and  final MSE, respectively.   

\subsection{Data collection}
We obtained the study approval from IISER Bhopal's ethics committee. IISER Bhopal students, regardless of their mental health status, were invited to fill out a Google Form indicating their preferred date and time for the study participation. They then received an email from a research assistant (RA) confirming their attendance at the venue. Upon arrival, participants received a participant information sheet and an informed consent form. After signing the consent form, they completed the MSE questionnaire in English, which took 20-25 minutes on average. A total of 405 participants (271 males and 134 females) participated over 120 days. Participant demographics are in Table \ref{tab:particiapnt_demographic}. After completing the study, participants were provided snacks to acknowledge their valuable time.

\begin{table}
    \centering
    \small
    
    \begin{tabular}{ccccc}
    \toprule
 & \textbf{\#} & \textbf{Age} & \textbf{Home Residence}\\
 & & ($\mu$, $\sigma$)& (urban, rural) &\\ \midrule
         All&  405&  (21.48, 3.59)&  (289, 116)\\
         Male&  271&  (21.17, 3.54)&  (189, 82)\\
         Female&  134&  (22.13, 3.62)&  (100, 34)\\ \bottomrule
    \end{tabular}
    \caption{Participants Demographics  }
    \label{tab:particiapnt_demographic}
\end{table}

\subsection{Dialogue representation}

We developed a Python script to transform participants' MSE questionnaire responses into simulated doctor-patient conversations to replicate real-world conversations. This process generated 405 doctor-patient conversation sessions, with 4860 (=  12 responses x 405 participants) utterances from participants and an equal number from doctors, totaling 9720 utterances. An anonymized excerpt of such a conversation for one participant is presented in Table \ref{tab:psychiatrist_patient_diaogue} in the appendix. Figure \ref{fig:question_answer_length} in the appendix shows the average length of utterances for each of the 12 questions. The average length of the dialogue conversation with and without the questionnaire is 3591 and 1987 characters.

\subsection{Reference human summaries}
To facilitate the training of supervised deep-learning models for summarizing doctor-patient conversations, reference summaries are required. Such summaries should encompass essential information, context, and insights of collected MSEs. Due to the lack of standardized guidelines for creating such summaries and the subjective nature of human-generated summaries influenced by personal perception, we developed a structured summary template similar to \citep{can2023uetcorn}.
Furthermore, given the structured nature of the MSE questions, the template was well-suited for summarization purposes. The summary template underwent thorough scrutiny through a rigorous review process involving feedback from three independent reviewers (i.e., graduate researchers). Subsequent revisions were made based on their input, ensuring the summary effectively captured key information while maintaining conciseness, clarity, and correctness.
After multiple iterations, the final version of the summary template was approved for use by a psychiatrist, leveraging their domain-specific knowledge. The template utilized to generate the reference summaries can be found in \ref{appendix: summary_template} in the appendix. The generated reference summary was further evaluated independently by five reviewers, as discussed in \ref{appendix: human_generated_summary_evaluation} in the appendix.


\subsection{Training}

To efficiently summarize MSE, we utilized language models designed for \textit{summarization}. Our dataset comprises simulated doctor-patient dialogues and human-generated reference summaries, making it suitable for supervised learning methods. 
 To our knowledge, no existing models publicly exist explicitly to summarize conversational psychological data. Rather than creating new models for our task, we opted to fine-tune existing summarization models, aligning with recent research trends in summarization ~~\citep{tang2023gersteinlab,mathur2023summqa,milintsevich2023calvados,feng2023improving}. We employed five models: BART-base, BART-large-CNN, T5-large, BART-large-xsum-samsum, and Pegasus-large~\citep{lewis2020bart,raffel2020exploring,gliwa2019samsum,zhang2020pegasus}.   As explained below, we chose these models over other available models for our task due to their appropriateness for the summarization task. 

 \begin{itemize}[leftmargin=*] \itemsep0em
 
    \item \textbf{BART base model} \citep{lewis2020bart}: It is a transformer encoder-decoder model featuring a bidirectional encoder and an autoregressive decoder.
    It demonstrates superior efficacy when fine-tuned for text-generation tasks such as summarization and translation~\citep{huang2020have}. In our evaluation, we utilized the BART base model from Hugging Face\footnote{\url{https://huggingface.co/facebook/bart-base}}, comprising 139 million parameters. 
    \item \textbf{BART-large-CNN model}: It is a fine-tuned model of BART-base with the CNN Daily Mail dataset~\citep{hermann2015teaching}. It is tailored for text summarization, leveraging a dataset containing a vast collection of articles, each accompanied by its summary. Given that the primary objective of BART-large-CNN is text summarization, we used it's Hugging Face\footnote{\url{https://huggingface.co/facebook/bart-large-cnn}} implementation, which has 406 million parameters.
    \item \textbf{T5 large}: The ``T5 Large for medical text summarization'' model is a tailored version of the T5 transformer model~\citep{raffel2020exploring}, fine-tuned to excel in summarizing medical text. It is fine-tuned on the dataset, encompassing a variety of medical documents, clinical studies, and healthcare research materials supplemented by human-generated summaries.
    Given that the model is designed for medical summarization tasks, we found it appropriate for fine-tuning on our psychological conversations. We used the model from Hugging Face\footnote{\url{https://huggingface.co/Falconsai/medical_summarization}}, which encompasses 60.5 million parameters.
    \item \textbf{BART-large-xsum-samsum model} \citep{gliwa2019samsum}: It is trained on the Samsum corpus dataset, comprising 16,369 conversations along with their respective summaries. Given that this model is explicitly trained on conversation data, it was deemed suitable for our task. We utilized the pre-trained model from Hugging Face\footnote{\url{https://huggingface.co/lidiya/bart-large-xsum-samsum}}, which contains 406 million parameters.
    \item \textbf{Pegasus-large} \citep{zhang2020pegasus}: It is a sequence-to-sequence model with an architecture similar to BART. However, it is pre-trained using two self-supervised objective functions: Masked Language Modeling \& a unique summarization-specific pre-training objective known as Gap Sentence Generation. We selected it because our input summary template also contains gaps, \& we wanted to assess its effectiveness in filling gaps while generating summaries. For this study, we used the pre-trained Pegasus large model with 568 million parameters from Hugging Face\footnote{\url{https://huggingface.co/google/pegasus-large}}.
\end{itemize}

Despite the significant progress in language models, training and fine-tuning them remains computationally intensive. Additionally, these models require high-performance computational resources to function effectively even after fine-tuning. Hence, we avoided using large language models such as Mistral, MentaLLaMA, and MentalBERT, which have billions of parameters~\citep{jiang2023mistral,yang2023mentalllama,ji2022mentalbert}. Their computational demands make them impractical for real-world applications, where systems typically have limited processing power and memory (around 16-32 GB of RAM). Our results demonstrate that billion-parameter models are unnecessary for our summarization task. Furthermore, considering the ethical and privacy concerns inherent in mental health care, we refrained from using online models like GPT-4. Instead, we prioritized offline-capable language models that can operate on standard home systems.

\section{Experiments}
We adopted the well-known ROUGE (Recall-Oriented Understudy for Gisting Evaluation) metric~\citep{lin2004rouge} as the primary evaluation criterion, in line with recent literature~\citep{krishna2021generating,zhang2021leveraging,michalopoulos2022medicalsum}. The metric compares the automated summary generated from the trained model with the reference summary. However, ROUGE scores have limitations, particularly in capturing factual consistency with the input text. Summary inconsistencies can range from inversions (e.g., negation) to incorrect usage of entities (e.g., subject-object swapping) or even hallucinations (e.g., introducing entities not present in the original document)~\citep{laban2022summac}. Recent studies have shown that even state of the art pre-trained language models can produce inconsistent summaries in over 70\% of specific scenarios~\citep{pagnoni2021understanding}. Hence, we also assessed the SummaC (Summary Consistency) score~\citep{laban2022summac} alongside ROUGE. 

SummaC is focused on evaluating factual consistency in summarization. It detects inconsistencies by splitting the reference and generated summaries into sentences and computing the entailment probabilities on all sentence pairs, where the premise is a reference summary sentence and the hypothesis is a generated summary sentence. It aggregates the SummaC scores for all pairs by training a convolutional neural network to aggregate the scores~\citep{laban2022summac}. We use the publicly available implementation\footnote{\url{https://github.com/tingofurro/summac}} for computing SummaC.

While these metrics excel at syntactical textual similarities, they fail to capture semantic similarities between two summaries. However, to address the limitation of the metric in terms of semantic analysis, we did qualitative analysis using ratings from clinical and non-clinical annotators to check the semantic similarities between reference and model-generated summaries. Additionally, we employed Large Language Models (LLMs) to evaluate the generated summaries.

The dataset comprising 405 conversations was divided into 270 for training, 68 for validation, and 67 for testing. The Appendix \ref{appendix: hyperparameters} lists the training settings, including hyperparameter settings utilized during model training. 

\begin{table*}[h]
\centering
{%
\begin{tabular}{lccccc}
\toprule
\textbf{Models}  &\textbf{Epochs (\#)}& \textbf{ROUGE-1}& \textbf{ROUGE-2}& \textbf{ROUGE-L}& \textbf{SummaC}\\ \midrule
BART-base  &25& 0.806& 0.686& 0.758 & 0.643\\ 
BART-large-CNN   &25& 0.815& 0.693& 0.774 & 0.714\\  
T5 large  &100& 0.752& 0.617& 0.697 & 0.545\\  
BART-large-xsum-samsum  &25& 0.804& 0.691& 0.764 & \textbf{0.724}\\ 
Pegasus-large &50& \textbf{0.829}& \textbf{0.710}& \textbf{0.790} & 0.699\\ \bottomrule
\end{tabular}%
}
\caption{ROUGE \& SummaC values of the model generated summaries with fine-tuning. Reported values represent the average values over the test set summaries of 67 doctor-patient conversations. \textbf{Higher ROUGE \& SummaC values indicate better summaries}.}
\label{tab:roUge_metric}
\end{table*}


\subsection{Quantitative evaluation}
The average ROUGE values (ROUGE-1, ROUGE-2, ROUGE-L,) and SummaC for the generated test set summaries with different models without and with fine-tuning are shown in Tables \ref{tab: appendix_roUge_metric} (appendix) and \ref{tab:roUge_metric} respectively. The values were computed by comparing the model generated and human reference summaries. 

Table~\ref{tab: appendix_roUge_metric} (appendix) shows that the BART-large-xsum-samsum model, without fine-tuning, attains the highest ROUGE across all mentioned ROUGE metrics, but the BART-base model achieves the highest SummaC. The low ROUGE and SummaC indicate that these models are not suitable for direct application in summarizing mental health conversation data. Moreover, after analyzing the output summaries generated by these models, we found that the pre-trained weights of these models tended to produce incomplete summaries, although they were able to capture smaller contexts of the conversation, as shown in Table  \ref{tab:model_varying_epochs} in the Appendix.

Following fine-tuning with our dataset, Pegasus-large achieved the highest ROUGE metric scores of 0.829, 0.710, and 0.790 for ROUGE-1, ROUGE-2, and ROUGE-L, respectively (see Table \ref{tab:roUge_metric}). BART-large-xsum-samsum gives the highest SummaC score but performs poorly in the ROUGE score.

\textit{Conclusion:}  Based on the ROUGE and SummaC results, the fine-tuned  Pegasus-large and BART-large-CNN emerged as the best-performing models. Consequently, we utilized the summary generated by both BART-large-CNN and Pegasus-large models for further assessments in the subsequent evaluation sections. The BART-large-CNN model checkpoint at  $25^{th}$ epoch and Pegasus-large model checkpoint at  $50^{th}$ epoch are made available for research and practical use in the Hugging Face repository\footnote{\url{https://huggingface.co/SIR-Lab/MSE_Summarizer}}.

\subsection{Qualitative human evaluation} \label{section: Human evaluation}

To evaluate the semantic effectiveness of the generated summaries, we conducted a qualitative analysis wherein we provided both the raw conversations (i.e., 11 raw conversations) and the generated summaries (both Pegasus-large \& BART-large-CNN) to evaluators. This analysis aimed to address two questions: (i) How effectively did the models create summaries that were complete, fluent, \& free of hallucinations and contradictions? This aspect is referred to as \textit{coarse-grained} human evaluation, focusing on overall quality. (ii) How effectively did the models capture the factual information presented in the conversations? This aspect is termed \textit{fine-grained} human evaluation, as it delves into various aspects in detail. By categorizing our analysis into coarse-grained and fine-grained, we captured both the overarching quality and nuanced factual consistency of the generated summaries.

To conduct this assessment, we employed a randomization algorithm to select 11 test conversations, which represented 16\% of our test dataset. These conversations were paired with their corresponding summaries generated by both the models. Subsequently, we thoroughly examined these pairs to evaluate their effectiveness.

\subsubsection{Coarse-grained human evaluation}
We conducted a qualitative analysis with the assistance of two clinicians (psychiatrists) and ten non-clinicians (graduate students not part of the study). The selected conversations, along with the summaries generated by Pegasus-large and BART-large-CNN, were provided to the reviewers. Notably, the reviewers were unaware of which models generated the summaries during the evaluation. Reviewers were instructed to assess summaries on a 5-point scale based on several evaluation parameters. The parameters selected following a brief literature survey~\citep{zhang2021leveraging, yao2022d4} are: (i) \textit {\textbf{Completeness}}: Does the summary cover all relevant aspects of the conversation?, (ii) \textit{\textbf{Fluency}}: Is the summary well structured, free from awkward phrases, and grammatically correct?, (iii) \textit{\textbf{Hallucination}}: Does the summary contain any extra information that was not presented by the patient?, (iv) \textit{\textbf{Contradiction}}: Does the summary contradict with the information provided by the patient? 

 
\noindent\textit{\textbf{Findings}}:
Table \ref{tab:summary_evaluation_metric} presents the average scores from clinicians, non-clinicians, and a combined evaluation for all four parameters used to assess the generated summaries from the best-performing models, Pegasus-large and BART-large-CNN, on the test data. The differences in quality between the summaries generated by these models are negligible, suggesting that both models produce summaries that are as readable as those created by humans. However, on average, Pegasus-large outperformed BART-large-CNN across all human evaluation parameters. Surprisingly, both models exhibited minimal instances of hallucination, which is a common issue in language models. Additionally, we noted a slightly higher occurrence of contradictions compared to hallucinations, albeit at a minimal level on the Likert scale rating of 5. Furthermore, we observed a slight discrepancy between the evaluations from clinicians and non-clinicians, suggesting that clinicians may prefer summaries with more detailed psychological information.

\noindent{\textit{\textbf{Inter-rater agreement}}}: Inter-rater agreement or inter-rater reliability or inter-observer agreement, refers to the level of agreement between two or more raters when assessing the same data. It is often measured using statistical measures such as Cohen's kappa (ranges between -1 and 1)~\citep{mchugh2012interrater}. A value of -1 and 1 indicates complete disagreement and agreement, respectively.

We computed Cohen's Kappa separately for two clinical reviewers and ten non-clinical reviewers for the summaries generated by the best models. Our clinical reviewers achieved Cohen's Kappa coefficients of 0.25 and 0.19 for Pegasus-large and BART-large-CNN, respectively, indicating moderate agreement. Among non-clinical reviewers, the average Cohen's Kappa coefficients were 0.43 and 0.45 for Pegasus-large and BART-large-CNN, respectively, which is higher compared to clinicians.
The higher agreement among non-clinicians compared to clinicians can be explained by the following factors: (1) Subjective Judgments of Clinicians: Clinicians use their expertise and experience to interpret symptoms and make diagnostic decisions, which can introduce variability in their assessments.
(ii) Focus of Non-Clinicians: Non-clinicians evaluated the  summaries primarily based on overall content and general comprehension rather than the nuanced clinical details that clinicians might prioritize. Table \ref{tab:inter_rater_clinical} displays the Cohen's Kappa coefficients among clinicians, while Table \ref{tab:Cohen_non_clinical} in the appendix presents the Cohen's Kappa coefficients among non-clinical reviewers.

\begin{table*}
\small
\centering
\begin{tabular}{lccccc}
\toprule
\textbf{Reviewer} & \textbf{Fine-tuned model} &  \textbf{Completeness} &  \textbf{Fluency} &  \textbf{Hallucination} & \textbf{Contradiction} \\ 
& \textbf{summary}& \textbf{($\mu$, $\sigma$)} & \textbf{($\mu$, $\sigma$)} & \textbf{($\mu$, $\sigma$)} &\textbf{($\mu$, $\sigma$)} \\ \midrule

\textbf{Clinician + non-clinician} &  \textit{Pegasus-large}&  (4.56, 0.69)&  (4.53, 0.67)&  (1.37, 0.59)& (1.65, 0.82) \\
\textbf{combined} & \textit{BART-large-CNN}& (4.39, 0.67)& (4.45, 0.64)& (1.23, 0.47)&(1.60, 0.63) \\ \midrule

\multirow{2}{*}{\textbf{Only non-clinicians}}& \textit{Pegasus-large}& (4.65, 0.58)& (4.60, 0.56)& (1.35, 0.58)&(1.65, 0.83) \\ 
& \textit{BART-large-CNN}& (4.44, 0.59)& (4.47, 0.58)& (1.23, 0.48)&(1.60, 0.63) \\ \midrule

\multirow{2}{*}{\textbf{Only Clinicians}}&  \textit{Pegasus-large}&  (4.13, 0.99)&  (4.18, 1.00)&  (1.45, 0.67)& (1.59, 0.73) \\
&  \textit{BART-large-CNN }&  (4.13, 0.94)&  (4.36, 0.90)&  (1.22, 0.42)& (1.63, 0.65) \\ \midrule

\multirow{2}{*}{\textbf{LLMs}}&  \textit{Pegasus-large}&  (4.63, 0.49)&  (4.27, 0.76)&  (1.40, 0.66)& (1.54, 0.91) \\
& \textit{BART-large-CNN} &  (4.40, 0.73)&  (4.31, 0.64)&  (1.81, 1.00)& (1.68, 0.77)\\ \bottomrule
\end{tabular}
\caption{Human (clinician, non-clinician) and LLM evaluation scores on five parameters (i.e., Completeness, Fluency, Hallucination, Contradiction). For \textit{Completeness} and \textit{Fluency}, a rating closer to 5 indicates the best, whereas for \textit{Hallucination} and \textit{Contradiction}, a rating closer to 1 is preferable.}
\label{tab:summary_evaluation_metric}
\end{table*}

\subsubsection{Fine-grained human evaluation}

To assess the factual consistency of the summaries, we engaged 10 graduate students who had previously participated in the coarse-grained human evaluation. These reviewers were provided with the conversation transcripts, model-generated summaries, and a questionnaire. The questionnaire consisted of two questions for each of eight \textit{parameters}: gender, mood, social life, academic pressure, concentration ability, difficulty with memory, strategies to feel better, and mental disorders. Reviewers were asked to respond with either ``Yes'' or ``No'' to the following questions for each parameter: (a) Does the summary capture the <parameter> of the input patient/participant conversation? (b) Is the summary data consistent with the provided conversation? Each evaluator had to answer 16 items on the questionnaire, providing a binary assessment for each parameter.
  
\noindent\textit{\textbf{Findings}}: 
Figure \ref{fig:HE_question_wise} shows the percentage of the parameters captured by our best-fine-tuned models on 11 test samples. The comprehensive analysis reveals that Pegasus-generated summaries captured parameters 92.8\% of the time, slightly surpassing BART-large-CNN’s coverage at 91.7\%. However, when analyzed by questionnaire sections (i.e., (a) and (b) as defined above), Pegasus-generated summaries (see Figure \ref{fig: HE_Pegasus_A} and \ref{fig: HE_Pegasus_B} in the appendix) show even higher accuracy, aligning with the conversation 98.4\% and 87.2\% of the time, respectively. Similarly, BART-generated summaries (see Figure \ref{fig: HE_BART-LARGE_A} \& \ref{fig: HE_BART-LARGE_B}) show an accuracy of 96.9\% and 86.5\% for (a) and (b) questions, respectively. These results indicate a high level of accuracy achieved by both models, with Pegasus-generated summaries outperforming BART-large-CNN.

\subsection{LLM based evaluation}
 
In recent years, there has been an increasing reliance on large language models like ChatGPT for evaluation purposes alongside human evaluators \citep{wu2023large,li2024leveraging} due to their scalability. However, owing to the sensitivity and privacy concerns surrounding mental health data and in alignment with human evaluation practices, we restricted our evaluation to only the 11 test data points, mirroring human evaluation processes. To accomplish this, we employed prompt engineering techniques (prompt is given in Appendix \ref{appendix: prompt}), instructing ChatGPT 3.5\footnote{\url{https://chat.openai.com/}} and Claude\footnote{\url{https://claude.ai/chats}} to emulate individuals proficient in the English language. Then, these large large language models were tasked to rate the summaries generated by Pegasus-large and BART-large-CNN based on original conversation data and to verify the factual consistency of the summaries.  We opted for the free versions of ChatGPT and Claude for this purpose.

 Table \ref{tab:summary_evaluation_metric} displays the average ratings acquired for completeness, fluency, hallucination, and contradiction in the summaries generated by Pegasus-large and BART-large-CNN. Meanwhile, Figures \ref{fig: llm_evaluation_plot} illustrate the percentage of parameters (gender, mood, social life, academic pressure, concentration ability, difficulty with memory, strategies to feel better, and mental disorders) captured by these models. According to the evaluation based on large language models, Pegasus-generated summaries captured parameters 85\% of the time, compared to BART-large-CNN's 83\%. This suggests that our fine-tuned model can generate summaries with moderately good evaluation parameters and a high percentage of parameters stated in the psychological conversation.

\section{Generalization}

To assess the generalizability of our two best fine-tuned models, we utilized the publicly available D4 dataset released by \citep{yao2022d4} and Emotional-Support-Conversation (ESC) dataset by Liu et al. \citep{liu2021towards}. Both D4 and  ESC data include a psychological conversation between a psychologist and a patient. We used five independent non-clinical reviewers (not part of our dataset summary evaluation) to rate the generated summaries of ten randomly selected conversations from the D4 and ESC. The parameters utilized for evaluating the generated summaries included \textit{completeness, fluency, hallucination,} and \textit{contradiction}, discussed previously in Section \ref{section: Human evaluation}.

Upon reviewing the reviewers' ratings, we found that the fine-tuned BART-large-CNN model's summary scored well in all parameters, as shown in Table \ref{tab:generalizability_evaluator_Score}. However, the performance of the fine-tuned Pegasus-large model's generated summary was notably poor, suggesting that our fine-tuned Pegasus-large model cannot be generalized. Table \ref{tab:chinese_dialogue_1} and \ref{tab:ESC} in the appendix presents dialogue conversations taken from \citep{yao2022d4} and \citep{liu2021towards}, respectively, alongside the corresponding summaries generated by the fine-tuned Pegasus-large and BART-large-CNN models.

\noindent{\textit{\textbf{Key finding}}}: 
While we noticed similar performance between BART-large-CNN and Pegasus-large on our dataset, there was a distinction in the case of these unseen data: Pegasus-large exhibited poor performance when applied to unseen data, whereas BART-large-CNN performed well with these unseen data. This suggests that our fine-tuned BART-large-CNN model demonstrates versatility, potentially capable of effectively processing psychological conversation datasets with good fluency and completeness while minimizing hallucination and contradictions.

\section{Implications of our study}


In this work, we presented the best-fine-tuned summarization models for generating accurate and concise summaries from MSEs for the attending doctor. 
The primary intention of this technology is not to replace doctors but to serve as an assistant to attending doctors by offering concise summaries of patients' mental health. This approach holds particular promise for implementation in low-income countries with a shortage of mental health professionals. However, further research is necessary to address privacy concerns and ensure the accuracy of the data utilized. The in-depth discussion can be found in section \ref{appendix: discussion} in the appendix.

In real-world scenarios, mental health service providers often lack access to such high-end systems, thereby limiting the practical application of language models in these settings. Our fine-tuned language models are tailored for specific tasks, i.e., summarization, and consist of 460 million and 568 million parameters for BART-large-CNN and Pegasus-large, respectively. We conducted experiments to assess the deployment of our language models on low-end systems without GPUs, and the results (shown in Table \ref{tab:low_end}) indicate that our fine-tuned models can operate effectively on such systems, providing reasonable response time.

\section{Conclusion}
The automatic generation of medical summaries from psychological patient conversations faces several challenges, including limited availability of publicly available data, significant domain shift from the typical pre-training text for transformer models, and unstructured lengthy dialogues. This paper investigates the potential of using pre-trained transformer models to summarize psychological patient conversations. We demonstrate that we can generate fluent and adequate summaries even with limited training data by fine-tuning transformer models on a specific dataset. Our resulting models outperform the performance of pre-trained models and surpass the quality of previously published work on this task. We evaluate transformer models for handling psychological conversations, compare pre-trained models with fine-tuned ones, and conduct extensive and intensive evaluations.

\section{Ethical considerations of our study}
Indeed, our psychological conversation data contain sensitive personal information about the participants and their experience. Therefore, we utilized anonymized numerical identifiers to store the participants' data for storage and further use. We ensured that the personal participants' information, such as name, age, and email address, could not be traced back using the anonymized numerical identifiers. Additionally, this study was approved by the ethics committee of the host institute.

 Although our experiments on fine-tuning summarization models have shown promising capabilities for summarizing conversation data, there is still a long way to go before they can be deployed in real-life systems. Recent research has revealed potential biases or harmful suggestions generated by language models \citep{xu2024mental}. Algorithms may reproduce or amplify societal biases in the training data, resulting in biased responses, recommendations, or the reinforcement of harmful narratives~\citep{mitchell2019model}. Biases may arise from limited training data that lack cultural and socioeconomic diversity, significantly affecting the usefulness of these models within the context of psychological counseling. Meanwhile, our study highlights the risks of hallucination, factual inconsistency, and contradiction in current language models. 
 
 Recent studies call for more research emphasis and efforts in assessing and mitigating these biases for mental health~\citep{chung2023challenges}.
 The black box nature of AI, i.e., the lack of interpretability of language models, poses significant challenges for their usage in psychological counseling. 
 Interpreting how these models process and generate responses becomes challenging, hindering transparency and accountability~\citep{ribeiro2020beyond}. The lack of interpretability also raises concerns regarding their use in the psychological domain.

 Privacy is another critical concern. However, addressing the challenges related to data security and patient privacy is paramount. By implementing appropriate data protection measures, ensuring patient consent, and adhering to ethical considerations, we can harness the potential of language models while safeguarding patient privacy.

\section{Limitations and Future Directions}

\begin{itemize}[leftmargin=*]
    \item When conducting MSE, it is important to note that MSE also encompasses the physical behavior \& appearance of the participants, which, we were unable to capture. However, this could be addressed by implementing a module where the front camera or webcam of participants' phones is activated while recording their responses.
    \item There were several instances where the participants' utterances were unclear to the reviewers. In real-world scenarios, when a patient's utterance is unclear, a doctor typically asks them to repeat and explain. However, in our case, this poses a major challenge. This issue could potentially be mitigated by testing the user's response for fluency and completeness after each utterance. If the model detects an issue, a new prompt could be sent to the user to encourage them to elaborate on their answers.
\end{itemize}

\balance
\bibliography{sample-base}

\begin{thebibliography}{59}
\providecommand{\natexlab}[1]{#1}

\bibitem[{Achiam et~al.(2023)Achiam, Adler, Agarwal, Ahmad, Akkaya, Aleman, Almeida, Altenschmidt, Altman, Anadkat et~al.}]{achiam2023gpt}
Josh Achiam, Steven Adler, Sandhini Agarwal, Lama Ahmad, Ilge Akkaya, Florencia~Leoni Aleman, Diogo Almeida, Janko Altenschmidt, Sam Altman, Shyamal Anadkat, et~al. 2023.
\newblock Gpt-4 technical report.
\newblock \emph{arXiv preprint arXiv:2303.08774}.

\bibitem[{Anthropic(2023)}]{anthropic}
Anthropic. 2023.
\newblock Introducing claude.

\bibitem[{Cai et~al.(2023)Cai, Wang, Wang, de~Melo, Zhang, Wang, and He}]{cai2023medbench}
Yan Cai, Linlin Wang, Ye~Wang, Gerard de~Melo, Ya~Zhang, Yanfeng Wang, and Liang He. 2023.
\newblock Medbench: A large-scale chinese benchmark for evaluating medical large language models.
\newblock \emph{arXiv preprint arXiv:2312.12806}.

\bibitem[{Can et~al.(2023)Can, Nguyen, Nguyen, Nguyen, Nguyen, Do, and Le}]{can2023uetcorn}
Duy-Cat Can, Quoc-An Nguyen, Binh-Nguyen Nguyen, Minh-Quang Nguyen, Khanh-Vinh Nguyen, Trung-Hieu Do, and Hoang-Quynh Le. 2023.
\newblock Uetcorn at mediqa-sum 2023: Template-based summarization for clinical note generation from doctor-patient conversation.
\newblock In \emph{CLEF}.

\bibitem[{Chen et~al.(2023)Chen, Gao, and He}]{chen2023evaluating}
Shiqi Chen, Siyang Gao, and Junxian He. 2023.
\newblock Evaluating factual consistency of summaries with large language models.
\newblock \emph{arXiv preprint arXiv:2305.14069}.

\bibitem[{Chung et~al.(2023)Chung, Dyer, and Brocki}]{chung2023challenges}
Neo~Christopher Chung, George Dyer, and Lennart Brocki. 2023.
\newblock Challenges of large language models for mental health counseling.
\newblock \emph{arXiv preprint arXiv:2311.13857}.

\bibitem[{Dery et~al.(2024)Dery, Kolawole, Kagey, Smith, Neubig, and Talwalkar}]{dery2024everybody}
Lucio Dery, Steven Kolawole, Jean-Francois Kagey, Virginia Smith, Graham Neubig, and Ameet Talwalkar. 2024.
\newblock Everybody prune now: Structured pruning of llms with only forward passes.
\newblock \emph{arXiv preprint arXiv:2402.05406}.

\bibitem[{Dixit et~al.(2023)Dixit, Wang, and Chen}]{dixit2023improving}
Tanay Dixit, Fei Wang, and Muhao Chen. 2023.
\newblock Improving factuality of abstractive summarization without sacrificing summary quality.
\newblock \emph{arXiv preprint arXiv:2305.14981}.

\bibitem[{Enarvi et~al.(2020)Enarvi, Amoia, Teba, Delaney, Diehl, Hahn, Harris, McGrath, Pan, Pinto et~al.}]{enarvi2020generating}
Seppo Enarvi, Marilisa Amoia, Miguel Del-Agua Teba, Brian Delaney, Frank Diehl, Stefan Hahn, Kristina Harris, Liam McGrath, Yue Pan, Joel Pinto, et~al. 2020.
\newblock Generating medical reports from patient-doctor conversations using sequence-to-sequence models.
\newblock In \emph{Proceedings of the first workshop on natural language processing for medical conversations}, pages 22--30.

\bibitem[{Feng et~al.(2023)Feng, Fan, Liu, Lin, Yao, Wu, Huang, Li, and Ma}]{feng2023improving}
Huawen Feng, Yan Fan, Xiong Liu, Ting-En Lin, Zekun Yao, Yuchuan Wu, Fei Huang, Yongbin Li, and Qianli Ma. 2023.
\newblock Improving factual consistency of text summarization by adversarially decoupling comprehension and embellishment abilities of llms.
\newblock \emph{arXiv preprint arXiv:2310.19347}.

\bibitem[{Gliwa et~al.(2019)Gliwa, Mochol, Biesek, and Wawer}]{gliwa2019samsum}
Bogdan Gliwa, Iwona Mochol, Maciej Biesek, and Aleksander Wawer. 2019.
\newblock Samsum corpus: A human-annotated dialogue dataset for abstractive summarization.
\newblock \emph{EMNLP-IJCNLP 2019}, page~70.

\bibitem[{Gupta et~al.(2022)Gupta, Singh, Sarkar, and Batra}]{gupta2022development}
Snehil Gupta, Swarndeep Singh, Siddharth Sarkar, and Atul Batra. 2022.
\newblock Development and validation of the ethical challenges in clinical situations-questionnaire (eccs-q) by involving health-care providers from a tertiary care health setting.
\newblock \emph{Clinical Ethics}, 17(2):172--183.

\bibitem[{Hermann et~al.(2015)Hermann, Kocisky, Grefenstette, Espeholt, Kay, Suleyman, and Blunsom}]{hermann2015teaching}
Karl~Moritz Hermann, Tomas Kocisky, Edward Grefenstette, Lasse Espeholt, Will Kay, Mustafa Suleyman, and Phil Blunsom. 2015.
\newblock Teaching machines to read and comprehend.
\newblock \emph{Advances in neural information processing systems}, 28.

\bibitem[{Huang et~al.(2020)Huang, Cui, Yang, Bao, Wang, Xie, and Zhang}]{huang2020have}
Dandan Huang, Leyang Cui, Sen Yang, Guangsheng Bao, Kun Wang, Jun Xie, and Yue Zhang. 2020.
\newblock What have we achieved on text summarization?
\newblock In \emph{Proceedings of the 2020 Conference on Empirical Methods in Natural Language Processing (EMNLP)}, pages 446--469.

\bibitem[{Jain et~al.(2022)Jain, Jangra, Saha, and Jatowt}]{jain2022survey}
Raghav Jain, Anubhav Jangra, Sriparna Saha, and Adam Jatowt. 2022.
\newblock A survey on medical document summarization.
\newblock \emph{arXiv preprint arXiv:2212.01669}.

\bibitem[{Ji et~al.(2021)Ji, Zhang, Ansari, Fu, Tiwari, and Cambria}]{ji2021mentalbert}
Shaoxiong Ji, Tianlin Zhang, Luna Ansari, Jie Fu, Prayag Tiwari, and Erik Cambria. 2021.
\newblock Mentalbert: Publicly available pretrained language models for mental healthcare.
\newblock \emph{arXiv preprint arXiv:2110.15621}.

\bibitem[{Ji et~al.(2022)Ji, Zhang, Ansari, Fu, Tiwari, and Cambria}]{ji2022mentalbert}
Shaoxiong Ji, Tianlin Zhang, Luna Ansari, Jie Fu, Prayag Tiwari, and Erik Cambria. 2022.
\newblock Mentalbert: Publicly available pretrained language models for mental healthcare.
\newblock In \emph{Proceedings of the Thirteenth Language Resources and Evaluation Conference}, pages 7184--7190.

\bibitem[{Jiang et~al.(2023)Jiang, Sablayrolles, Mensch, Bamford, Chaplot, Casas, Bressand, Lengyel, Lample, Saulnier et~al.}]{jiang2023mistral}
Albert~Q Jiang, Alexandre Sablayrolles, Arthur Mensch, Chris Bamford, Devendra~Singh Chaplot, Diego de~las Casas, Florian Bressand, Gianna Lengyel, Guillaume Lample, Lucile Saulnier, et~al. 2023.
\newblock Mistral 7b.
\newblock \emph{arXiv preprint arXiv:2310.06825}.

\bibitem[{Jones and Hunter(1995)}]{jones1995consensus}
Jeremy Jones and Duncan Hunter. 1995.
\newblock Consensus methods for medical and health services research.
\newblock \emph{BMJ: British Medical Journal}, 311(7001):376.

\bibitem[{Joseph et~al.(2024)Joseph, Chen, Trienes, G{\"o}ke, Coers, Xu, Wallace, and Li}]{joseph2024factpico}
Sebastian~Antony Joseph, Lily Chen, Jan Trienes, Hannah~Louisa G{\"o}ke, Monika Coers, Wei Xu, Byron~C Wallace, and Junyi~Jessy Li. 2024.
\newblock Factpico: Factuality evaluation for plain language summarization of medical evidence.
\newblock \emph{arXiv preprint arXiv:2402.11456}.

\bibitem[{Krishna et~al.(2021)Krishna, Khosla, Bigham, and Lipton}]{krishna2021generating}
Kundan Krishna, Sopan Khosla, Jeffrey~P Bigham, and Zachary~C Lipton. 2021.
\newblock Generating soap notes from doctor-patient conversations using modular summarization techniques.
\newblock In \emph{Proceedings of the 59th Annual Meeting of the Association for Computational Linguistics and the 11th International Joint Conference on Natural Language Processing (Volume 1: Long Papers)}, pages 4958--4972.

\bibitem[{Kuzmin et~al.(2024)Kuzmin, Nagel, Van~Baalen, Behboodi, and Blankevoort}]{kuzmin2024pruning}
Andrey Kuzmin, Markus Nagel, Mart Van~Baalen, Arash Behboodi, and Tijmen Blankevoort. 2024.
\newblock Pruning vs quantization: Which is better?
\newblock \emph{Advances in Neural Information Processing Systems}, 36.

\bibitem[{Laban et~al.(2022)Laban, Schnabel, Bennett, and Hearst}]{laban2022summac}
Philippe Laban, Tobias Schnabel, Paul~N Bennett, and Marti~A Hearst. 2022.
\newblock Summac: Re-visiting nli-based models for inconsistency detection in summarization.
\newblock \emph{Transactions of the Association for Computational Linguistics}, 10:163--177.

\bibitem[{Lewis et~al.(2020)Lewis, Liu, Goyal, Ghazvininejad, Mohamed, Levy, Stoyanov, and Zettlemoyer}]{lewis2020bart}
Mike Lewis, Yinhan Liu, Naman Goyal, Marjan Ghazvininejad, Abdelrahman Mohamed, Omer Levy, Veselin Stoyanov, and Luke Zettlemoyer. 2020.
\newblock Bart: Denoising sequence-to-sequence pre-training for natural language generation, translation, and comprehension.
\newblock In \emph{Proceedings of the 58th Annual Meeting of the Association for Computational Linguistics}, pages 7871--7880.

\bibitem[{Li et~al.(2023)Li, Bubeck, Eldan, Del~Giorno, Gunasekar, and Lee}]{li2023textbooks}
Yuanzhi Li, S{\'e}bastien Bubeck, Ronen Eldan, Allie Del~Giorno, Suriya Gunasekar, and Yin~Tat Lee. 2023.
\newblock Textbooks are all you need ii: phi-1.5 technical report.
\newblock \emph{arXiv preprint arXiv:2309.05463}.

\bibitem[{Li et~al.(2024)Li, Xu, Shen, Xu, Gu, and Tao}]{li2024leveraging}
Zhen Li, Xiaohan Xu, Tao Shen, Can Xu, Jia-Chen Gu, and Chongyang Tao. 2024.
\newblock Leveraging large language models for nlg evaluation: A survey.
\newblock \emph{arXiv preprint arXiv:2401.07103}.

\bibitem[{Lin(2004)}]{lin2004rouge}
Chin-Yew Lin. 2004.
\newblock Rouge: A package for automatic evaluation of summaries.
\newblock In \emph{Text summarization branches out}, pages 74--81.

\bibitem[{Liu et~al.(2021)Liu, Zheng, Demasi, Sabour, Li, Yu, Jiang, and Huang}]{liu2021towards}
Siyang Liu, Chujie Zheng, Orianna Demasi, Sahand Sabour, Yu~Li, Zhou Yu, Yong Jiang, and Minlie Huang. 2021.
\newblock Towards emotional support dialog systems.
\newblock \emph{arXiv preprint arXiv:2106.01144}.

\bibitem[{Majumdar(2022)}]{majumdar2022covid}
Promita Majumdar. 2022.
\newblock Covid-19, unforeseen crises and the launch of national tele-mental health program in india.
\newblock \emph{Journal of Mental Health}, 31(4):451--452.

\bibitem[{Martin(1990)}]{martin1990mental}
David~C Martin. 1990.
\newblock The mental status examination.
\newblock \emph{Clinical Methods: The History, Physical, and Laboratory Examinations. 3rd edition}.

\bibitem[{Mathur et~al.(2023)Mathur, Rangreji, Kapoor, Palavalli, Bertsch, and Gormley}]{mathur2023summqa}
Yash Mathur, Sanketh Rangreji, Raghav Kapoor, Medha Palavalli, Amanda Bertsch, and Matthew~R Gormley. 2023.
\newblock Summqa at mediqa-chat 2023: in-context learning with gpt-4 for medical summarization.
\newblock \emph{arXiv preprint arXiv:2306.17384}.

\bibitem[{McHugh(2012)}]{mchugh2012interrater}
Mary~L McHugh. 2012.
\newblock Interrater reliability: the kappa statistic.
\newblock \emph{Biochemia medica}, 22(3):276--282.

\bibitem[{Michalopoulos et~al.(2022)Michalopoulos, Williams, Singh, and Lin}]{michalopoulos2022medicalsum}
George Michalopoulos, Kyle Williams, Gagandeep Singh, and Thomas Lin. 2022.
\newblock Medicalsum: A guided clinical abstractive summarization model for generating medical reports from patient-doctor conversations.
\newblock In \emph{Findings of the Association for Computational Linguistics: EMNLP 2022}, pages 4741--4749.

\bibitem[{Milintsevich and Agarwal(2023)}]{milintsevich2023calvados}
Kirill Milintsevich and Navneet Agarwal. 2023.
\newblock Calvados at mediqa-chat 2023: Improving clinical note generation with multi-task instruction finetuning.
\newblock In \emph{Proceedings of the 5th Clinical Natural Language Processing Workshop}, pages 529--535.

\bibitem[{Mitchell et~al.(2019)Mitchell, Wu, Zaldivar, Barnes, Vasserman, Hutchinson, Spitzer, Raji, and Gebru}]{mitchell2019model}
Margaret Mitchell, Simone Wu, Andrew Zaldivar, Parker Barnes, Lucy Vasserman, Ben Hutchinson, Elena Spitzer, Inioluwa~Deborah Raji, and Timnit Gebru. 2019.
\newblock Model cards for model reporting.
\newblock In \emph{Proceedings of the conference on fairness, accountability, and transparency}, pages 220--229.

\bibitem[{Pagnoni et~al.(2021)Pagnoni, Balachandran, and Tsvetkov}]{pagnoni2021understanding}
Artidoro Pagnoni, Vidhisha Balachandran, and Yulia Tsvetkov. 2021.
\newblock Understanding factuality in abstractive summarization with frank: A benchmark for factuality metrics.
\newblock \emph{arXiv preprint arXiv:2104.13346}.

\bibitem[{Qiu et~al.(2023)Qiu, He, Zhang, Li, and Lan}]{qiu2023smile}
Huachuan Qiu, Hongliang He, Shuai Zhang, Anqi Li, and Zhenzhong Lan. 2023.
\newblock Smile: Single-turn to multi-turn inclusive language expansion via chatgpt for mental health support.
\newblock \emph{arXiv preprint arXiv:2305.00450}.

\bibitem[{Radford et~al.(2018)Radford, Narasimhan, Salimans, Sutskever et~al.}]{radford2018improving}
Alec Radford, Karthik Narasimhan, Tim Salimans, Ilya Sutskever, et~al. 2018.
\newblock Improving language understanding by generative pre-training.

\bibitem[{Raffel et~al.(2020)Raffel, Shazeer, Roberts, Lee, Narang, Matena, Zhou, Li, and Liu}]{raffel2020exploring}
Colin Raffel, Noam Shazeer, Adam Roberts, Katherine Lee, Sharan Narang, Michael Matena, Yanqi Zhou, Wei Li, and Peter~J Liu. 2020.
\newblock Exploring the limits of transfer learning with a unified text-to-text transformer.
\newblock \emph{The Journal of Machine Learning Research}, 21(1):5485--5551.

\bibitem[{Ribeiro et~al.(2020)Ribeiro, Wu, Guestrin, and Singh}]{ribeiro2020beyond}
Marco~Tulio Ribeiro, Tongshuang Wu, Carlos Guestrin, and Sameer Singh. 2020.
\newblock Beyond accuracy: Behavioral testing of nlp models with checklist.
\newblock \emph{arXiv preprint arXiv:2005.04118}.

\bibitem[{Rojas et~al.(2019)Rojas, Mart{\'\i}nez, Mart{\'\i}nez, Franco, and Jim{\'e}nez-Molina}]{rojas2019improving}
Graciela Rojas, Vania Mart{\'\i}nez, Pablo Mart{\'\i}nez, Pamela Franco, and {\'A}lvaro Jim{\'e}nez-Molina. 2019.
\newblock Improving mental health care in developing countries through digital technologies: a mini narrative review of the chilean case.
\newblock \emph{Frontiers in public health}, 7:391.

\bibitem[{Saraceno et~al.(2007)Saraceno, van Ommeren, Batniji, Cohen, Gureje, Mahoney, Sridhar, and Underhill}]{saraceno2007barriers}
Benedetto Saraceno, Mark van Ommeren, Rajaie Batniji, Alex Cohen, Oye Gureje, John Mahoney, Devi Sridhar, and Chris Underhill. 2007.
\newblock Barriers to improvement of mental health services in low-income and middle-income countries.
\newblock \emph{The Lancet}, 370(9593):1164--1174.

\bibitem[{Song et~al.(2024)Song, Chim, Tsakalidis, Ive, Atzil-Slonim, and Liakata}]{song2024clinically}
Jiayu Song, Jenny Chim, Adam Tsakalidis, Julia Ive, Dana Atzil-Slonim, and Maria Liakata. 2024.
\newblock Clinically meaningful timeline summarisation in social media for mental health monitoring.
\newblock \emph{arXiv preprint arXiv:2401.16240}.

\bibitem[{Song et~al.(2020)Song, Tian, Wang, and Xia}]{song2020summarizing}
Yan Song, Yuanhe Tian, Nan Wang, and Fei Xia. 2020.
\newblock Summarizing medical conversations via identifying important utterances.
\newblock In \emph{Proceedings of the 28th International Conference on Computational Linguistics}, pages 717--729.

\bibitem[{Srivastava et~al.(2022)Srivastava, Suresh, Lord, Akhtar, and Chakraborty}]{srivastava2022counseling}
Aseem Srivastava, Tharun Suresh, Sarah~P Lord, Md~Shad Akhtar, and Tanmoy Chakraborty. 2022.
\newblock Counseling summarization using mental health knowledge guided utterance filtering.
\newblock In \emph{Proceedings of the 28th ACM SIGKDD Conference on Knowledge Discovery and Data Mining}, pages 3920--3930.

\bibitem[{Tang et~al.(2023)Tang, Tran, Tan, and Gerstein}]{tang2023gersteinlab}
Xiangru Tang, Andrew Tran, Jeffrey Tan, and Mark Gerstein. 2023.
\newblock Gersteinlab at mediqa-chat 2023: Clinical note summarization from doctor-patient conversations through fine-tuning and in-context learning.
\newblock \emph{arXiv preprint arXiv:2305.05001}.

\bibitem[{Touvron et~al.(2023)Touvron, Lavril, Izacard, Martinet, Lachaux, Lacroix, Rozi{\`e}re, Goyal, Hambro, Azhar et~al.}]{touvron2023llama}
Hugo Touvron, Thibaut Lavril, Gautier Izacard, Xavier Martinet, Marie-Anne Lachaux, Timoth{\'e}e Lacroix, Baptiste Rozi{\`e}re, Naman Goyal, Eric Hambro, Faisal Azhar, et~al. 2023.
\newblock Llama: Open and efficient foundation language models.
\newblock \emph{arXiv preprint arXiv:2302.13971}.

\bibitem[{Voss et~al.(2019)}]{voss2019mental}
Rachel~M Voss et~al. 2019.
\newblock Mental status examination.

\bibitem[{WHO(2022)}]{world2022world}
World Health~Organization WHO. 2022.
\newblock World mental health report: transforming mental health for all.

\bibitem[{Wu et~al.(2023)Wu, Gong, Shou, Liang, and Jiang}]{wu2023large}
Ning Wu, Ming Gong, Linjun Shou, Shining Liang, and Daxin Jiang. 2023.
\newblock Large language models are diverse role-players for summarization evaluation.
\newblock In \emph{CCF International Conference on Natural Language Processing and Chinese Computing}, pages 695--707. Springer.

\bibitem[{Xu et~al.(2023)Xu, Yao, Dong, Yu, Hendler, Dey, and Wang}]{xu2023leveraging}
Xuhai Xu, Bingshen Yao, Yuanzhe Dong, Hong Yu, James Hendler, Anind~K Dey, and Dakuo Wang. 2023.
\newblock Leveraging large language models for mental health prediction via online text data.
\newblock \emph{arXiv preprint arXiv:2307.14385}.

\bibitem[{Xu et~al.(2024)Xu, Yao, Dong, Gabriel, Yu, Hendler, Ghassemi, Dey, and Wang}]{xu2024mental}
Xuhai Xu, Bingsheng Yao, Yuanzhe Dong, Saadia Gabriel, Hong Yu, James Hendler, Marzyeh Ghassemi, Anind~K Dey, and Dakuo Wang. 2024.
\newblock Mental-llm: Leveraging large language models for mental health prediction via online text data.
\newblock \emph{Proceedings of the ACM on Interactive, Mobile, Wearable and Ubiquitous Technologies}, 8(1):1--32.

\bibitem[{Yang et~al.(2023)Yang, Zhang, Kuang, Xie, and Ananiadou}]{yang2023mentalllama}
Kailai Yang, Tianlin Zhang, Ziyan Kuang, Qianqian Xie, and Sophia Ananiadou. 2023.
\newblock Mentalllama: Interpretable mental health analysis on social media with large language models.
\newblock \emph{arXiv preprint arXiv:2309.13567}.

\bibitem[{Yao et~al.(2022)Yao, Shi, Zou, Dai, Wu, Chen, Wang, and Yu}]{yao2022d4}
Binwei Yao, Chao Shi, Likai Zou, Lingfeng Dai, Mengyue Wu, Lu~Chen, Zhen Wang, and Kai Yu. 2022.
\newblock D4: a chinese dialogue dataset for depression-diagnosis-oriented chat.
\newblock In \emph{Proceedings of the 2022 Conference on Empirical Methods in Natural Language Processing}, pages 2438--2459.

\bibitem[{Yun et~al.(2023)Yun, Sohn, and Kyeong}]{yun2023fine}
Jiseon Yun, Jae~Eui Sohn, and Sunghyon Kyeong. 2023.
\newblock Fine-tuning pretrained language models to enhance dialogue summarization in customer service centers.
\newblock In \emph{Proceedings of the Fourth ACM International Conference on AI in Finance}, pages 365--373.

\bibitem[{Zablotskaia et~al.(2023)Zablotskaia, Khalman, Joshi, Soares, Jakobovits, Maynez, and Narayan}]{zablotskaia2023calibrating}
Polina Zablotskaia, Misha Khalman, Rishabh Joshi, Livio~Baldini Soares, Shoshana Jakobovits, Joshua Maynez, and Shashi Narayan. 2023.
\newblock Calibrating likelihoods towards consistency in summarization models.
\newblock \emph{arXiv preprint arXiv:2310.08764}.

\bibitem[{Zhang et~al.(2020)Zhang, Zhao, Saleh, and Liu}]{zhang2020pegasus}
Jingqing Zhang, Yao Zhao, Mohammad Saleh, and Peter Liu. 2020.
\newblock Pegasus: Pre-training with extracted gap-sentences for abstractive summarization.
\newblock In \emph{International conference on machine learning}, pages 11328--11339. PMLR.

\bibitem[{Zhang et~al.(2021)Zhang, Negrinho, Ghosh, Jagannathan, Hassanzadeh, Schaaf, and Gormley}]{zhang2021leveraging}
Longxiang Zhang, Renato Negrinho, Arindam Ghosh, Vasudevan Jagannathan, Hamid~Reza Hassanzadeh, Thomas Schaaf, and Matthew~R Gormley. 2021.
\newblock Leveraging pretrained models for automatic summarization of doctor-patient conversations.
\newblock In \emph{Findings of the Association for Computational Linguistics: EMNLP 2021}, pages 3693--3712.

\bibitem[{Zhong et~al.(2022)Zhong, Liu, Xu, Zhu, and Zeng}]{zhong2022dialoglm}
Ming Zhong, Yang Liu, Yichong Xu, Chenguang Zhu, and Michael Zeng. 2022.
\newblock Dialoglm: Pre-trained model for long dialogue understanding and summarization.
\newblock In \emph{Proceedings of the AAAI Conference on Artificial Intelligence}, volume~36, pages 11765--11773.

\end{thebibliography}

\clearpage

\appendix

\onecolumn
\section{Appendix}

\setcounter{table}{0}
\setcounter{figure}{0}
\renewcommand{\thetable}{A.\arabic{table}}
\renewcommand{\thefigure}{A.\arabic{figure}}

\subsection{MSE Questionnaire}

\begin{table}[H]
    \centering
    \small
    \begin{tabular}{p{15.5cm}}
    \toprule 
Q1. Please describe your social life at the college campus. Are you actively participating in extracurricular activities, interacting with others, or taking initiative to socialize with others?\\
Q2. Describe your typical daily mood?\\
Q3. Does your mood remain steady or goes up and down throughout the day without any reason or on trivial matters?\\
Q4. How do you handle day-to-day irritations or frustrations? \\
Q5. How do you handle pressure related to academics?\\
Q6. Describe your ability to attend to the task at hand or concentrate on daily tasks (academic, non-academic)? \\
Q7. Have you noticed any difficulties with memory, such as unable to register new information, forgetting recent events, or not able to recall older personal/factual events? \\
Q8. What do you do to feel better? For example, some people take caffeine, talk with people, or watch movies to feel better.\\
Q9. Describe how supported you feel by others (e.g., friends, family) around you and how they help you? \\
Q10. What do you usually do when you have a bad day or when you are not able to concentrate on work? \\
Q11. Are you experiencing symptoms of stress, anxiety, or depression? If yes, describe the symptoms?\\
Q12. Are you doing anything (by self or help seeking) for the ongoing stress, anxiety, or depression, if any? If yes, what?\\ \bottomrule 
    \end{tabular}
    \caption{Final MSE questionnaire}
    \label{tab:MSE Questionnaire}
\end{table}

\subsubsection{Questionnaire validation} \label{appendix: mse_questionnaire validation}
 
After finalizing the questionnaire, we conducted a survey with clinical psychiatrists. Initially, we introduced the MSE questionnaire developed by our team and presented the problem statement we aimed to address. Psychiatrists were then asked to evaluate the questionnaire based on item accuracy, language clarity, and reliability, following the guidelines outlined in the studies by Jones et al. \citep{jones1995consensus} and Gupta et al. \citep{gupta2022development}. They provided ratings on a scale from 1 (poor) to 5 (excellent). Four psychiatrists, not affiliated with the study team, participated in the survey. The average ratings obtained were 4.1 for item accuracy, 4.0 for language clarity, and 4.0 for reliability.
Subsequently, incorporating their feedback and suggestions, we finalized the questionnaire. The refined version is presented in Table \ref{tab:MSE Questionnaire} in the Appendix. Additionally, detailed average ratings per question are provided in Table \ref{tab:MSE Questionnaire evaluation score} of the appendix.

\begin{table*}[ht!]
    \centering
    \small
    \begin{tabular}{p{10cm}ccc}
    \toprule[0.5ex]
\multicolumn{1}{c}{\textbf{MSE Questions}}  &  Accuracy &  Language & Reliability  \\
\midrule
Q1. Please describe your social life at the college campus. Are you actively participating in extracurricular activities, interacting with others, or taking initiative to socialize with others? &    4.00
&    4.25
&    3.75
\\
Q2. Describe your typical daily Mood? &    3.75
&    4.00
&    3.50
\\
Q3. Does your Mood remain steady or goes up and down throughout the day without any reason or on trivial matters? &    3.75
&    3.50
&    4.00
\\
Q4. How do you handle day-to-day irritations or frustrations?  &    4.25
&    4.00
&    3.75
\\
Q5. How do you handle pressure related to academics? &    4.00
&    4.00
&    4.00
\\
Q6. Describe your ability to attend to the task at hand or concentrate on daily tasks (academic, non-academic)?  &    4.00
&    4.00
&    4.25
\\
Q7. Have you noticed any difficulties with memory, such as unable to register new information, forgetting recent events, or not able to recall older personal/factual events?  &    4.00
&    4.00
&    4.00
\\
Q8. What do you do to feel better? For example, some people take caffeine, talk with people, or watch movies to feel better. &    4.00
&    3.75
&    4.00
\\
Q9. Describe how supported you feel by others (e.g., friends, family) around you and how they help you?  &    4.25
&    4.25
&    4.25
\\
Q10. What do you usually do when you have a bad day or when you are not able to concentrate on work?  &    4.25
&    4.25
&    4.25
\\
Q11. Are you experiencing symptoms of stress, anxiety, or depression? If yes, describe the symptoms? &    4.25
&    3.75
&    4.00
\\
Q12. Are you doing anything (by self or help seeking) for the ongoing stress, anxiety, or depression, if any? If yes, what? &    4.25
&    3.75
&    4.25
\\ \midrule
\textbf{Average} & 4.06& 3.96&4.00 \\ \bottomrule[0.5ex]
    \end{tabular}
    \caption{Finalized MSE Questionnaire}
    \label{tab:MSE Questionnaire evaluation score}
\end{table*}

\subsection{Sample conversation}
\begin{table*}[ht]
\small
    \centering
    \begin{tabular}{>{\columncolor[HTML]{E8E8E8}} p{16cm } }
\textcolor{red}{Doctor}: \textit{What is your patient ID?}  \\
\textcolor{blue}{Patient}: \textit{\#3A4C307} \\
\textcolor{red}{Doctor}: \textit{What is your age?}  \\
\textcolor{blue}{Patient}:\textit{ 19} \\
\textcolor{red}{Doctor}: \textit{What is your gender?}  \\
\textcolor{blue}{Patient}: \textit{Female} \\
\textcolor{red}{Doctor}: \textit{Please describe your social life at the college campus. Are you actively participating in extracurricular activities, interacting with others, or taking initiative to socialize with others?}  \\ 
\textcolor{blue}{Patient}: \textit{I am part of football team and a core member in Physics club and Singularity working team. I also make contacts with my seniors and other staffs for both personal and official discussions. I volunteer for community fests and other initiatives.
} \\
\textcolor{red}{Doctor}:  \textit{Describe your typical daily Mood?}   \\
\textcolor{blue}{Patient}: \textit{I always try to find happiness in every single moment of my life. But at times I turnout t be moody.} \\
\textcolor{red}{Doctor}:  \textit{Does your Mood remain steady or goes up and down throughout the day without any reason or on trivial matters?} \\
\textcolor{blue}{Patient}: \textit{My mood is dynamic. It goes up and down for  both valid and unknown reasons. I get upset on simple jokes and responses from my close circle.} \\
\textcolor{red}{Doctor}: \textit{How do you handle day-to-day irritations or frustrations?}  \\
\textcolor{blue}{Patient}: \textit{I try to connect more with the Almighty through daily prayers. But mostly I prefer sleeping with no disturbance for hours. Nowadays I try to engage myself with a busy schedule and locations.} \\
\textcolor{red}{Doctor}: \textit{How do you handle pressure related to academics?}  \\
\textcolor{blue}{Patient}: \textit{lately I started purposeful ignorance of academic pressure. I will engage my times studying or with close friend. I also try to phone my parents when I feel so exhausted.} \\
\textcolor{red}{Doctor}: \textit{Describe your ability to attend to the task at hand or concentrate on daily tasks (academic, non-academic)?} \\
\textcolor{blue}{Patient}: \textit{I am mostly able to focus on my task and complete on time. But when I am in a bad mood I will distract myself from the task with social media and resume when I feel fine.} \\
\textcolor{red}{Doctor}: \textit{Have you noticed any difficulties with memory, such as unable to register new information, forgetting recent events, or not able to recall older personal/factual events?} \\
\textcolor{blue}{Patient}: \textit{Yes I do, and only very lately. I find it very difficult to comprehend what I see and try reading. I also noticed forgetting recent events which where not very important but still to be considered. I also have difficulty in recalling but the least.} \\
\textcolor{red}{Doctor}: \textit{What do you do to feel better? For example, some people take caffeine, talk with people, or watch movies to feel better.} \\
\textcolor{blue}{Patient}: \textit{Sleep mostly. But if it is with communication gap, I only settle after conveying my last note. I also sing a song or try dancing in my room but I prefer privacy for this} \\
\textcolor{red}{Doctor}: \textit{Describe how supported you feel by others (e.g., friends, family) around you and how they help you?}  \\
\textcolor{blue}{Patient}: \textit{I feel supported very less even from family. And so I don't expect any support from anyone and try to figure out all alone.} \\
\textcolor{red}{Doctor}: \textit{What do you usually do when you have a bad day or when you are not able to concentrate on work?}  \\
\textcolor{blue}{Patient}: \textit{I sleep for hours or the entire day. I also get some ease after crying or talking about it. I used talk to myself which helped me figure out the situation and motivated to push through.} \\
\textcolor{red}{Doctor}:  \textit{Are you experiencing symptoms of stress, anxiety, or depression? If yes, describe the symptoms?} \\
\textcolor{blue}{Patient}: \textit{Yes, all stress, anxiety and depression} \\
\textcolor{red}{Doctor}:  \textit{Are you doing anything (by self or help-seeking) for the ongoing stress, anxiety, or depression, if any? If yes, what?}  \\
\textcolor{blue}{Patient}: \textit{Yes, I'm reading books on self-development and self-improvement.} \\
    \end{tabular}
    \caption{Doctor-patient conversation dialogue of an anonymized participant.}
    \label{tab:psychiatrist_patient_diaogue}
\end{table*}

\begin{figure}[H]
    \centering
    \includegraphics[scale=0.5]{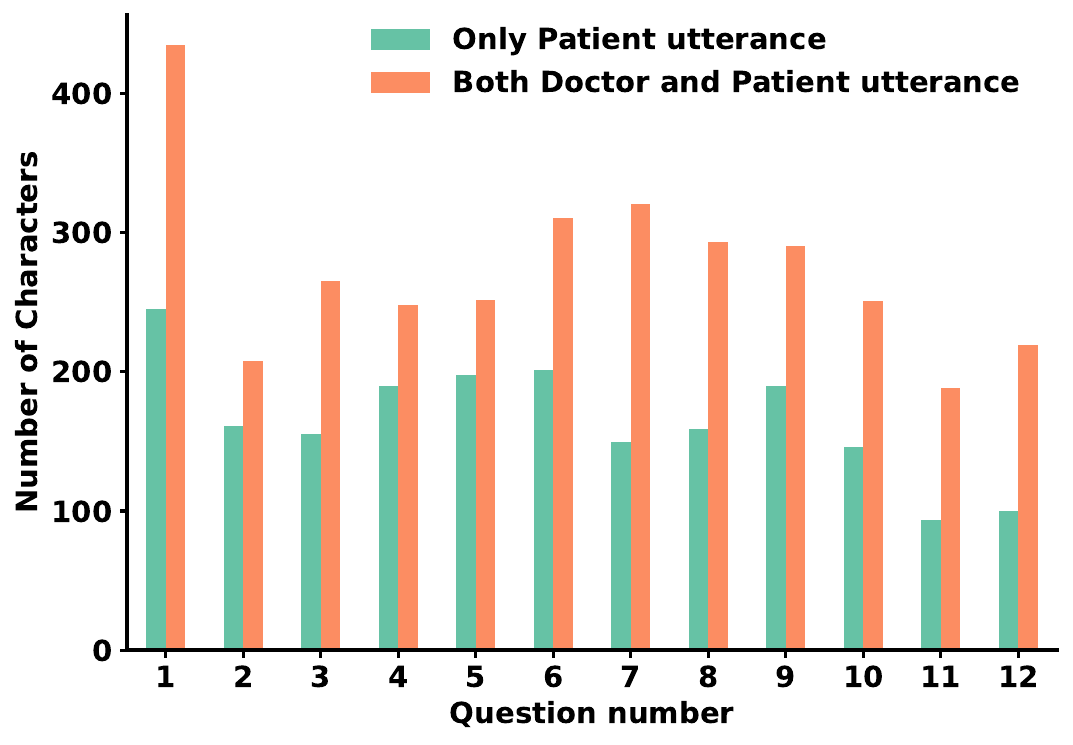}
    \caption{Average lengths of patient (i.e., participant) and doctor utterances for each question, aggregated across all 405 patient-doctor conversations. Note that the length of doctor utterances remains constant for each questionnaire, as the questions were predefined.}   
    \label{fig:question_answer_length}
\end{figure}

\subsection{Summary template} \label{appendix: summary_template}

 \begin{quote}
     Patient is a \rule{0.8cm}{0.15mm} year old [girl/boy/lady/man]. [His/Her] mood is generally \rule{0.8cm}{0.15mm} and [remains steady/but goes up and down] throughout the day. [He/She] [takes/does not take] part in extracurricular activities and  \rule{0.8cm}{0.15mm} [socializes/does not socialize] with others. For daily frustration  [He/She] does (*activities*). [He/She] [feels/does not feel] academic pressure and for this [He/She] does (*activities*). [His/Her] concentration and task attending ability is [good/bad]. [He/She] [feels/does not feel] difficulty with memory. [He/She] feels better by doing (*activities*).  [He/She] [feels/does not feel] supported by his family and friends. On a bad day, [he/she] prefers  \rule{0.8cm}{0.15mm}.  [He/She] is [experiencing/ not experiencing] \rule{0.8cm}{0.15mm}[stress/anxiety/depression] symptoms such as \rule{0.8cm}{0.15mm}.  
  \end{quote}

\subsubsection{Human generated summary evaluation} \label{appendix: human_generated_summary_evaluation}

To assess the template's efficacy in capturing the context of the MSE and user responses, we initially generated summaries (i.e., human-generated summaries) using the template with data from ten randomly selected participants. Subsequently, these summaries were evaluated based on completeness (i.e., whether the summary covers all relevant aspects of the conversation?) and Fluency (i.e., is the summary well structured, free from awkward phrases and grammatically?) on a scale of 1 (poor) to 5 (excellent).
The average ratings from 5 reviewers for each parameter were computed, revealing that the template effectively captured the MSE and user responses with a completeness rating of 4.66 and a fluency rating of 4.36.

\subsection{Training settings} \label{appendix: hyperparameters}

The models were trained on an NVIDIA A100-PCIE-40GB GPU, with an average training time of 2 hours per model. Our dataset consisted of 405 conversations, which we split into 270 for training, 68 for validation, and 67 for testing purposes. We conducted our experiments using varying numbers of epochs to evaluate the models' learning capabilities. Specifically, we trained the models for 5, 10, 25, 50, and 100 epochs.
Across all five models (BART-base, BART-large-CNN, T5 large, BART-large-xsum-samsum, and Pegasus), we maintained consistent hyperparameters using the PyTorch module with the following settings: \{\textit{max token length}: 1024 tokens,
     \textit{warmup steps}: 500,
     \textit{weight decay}: 0.01,
     \textit{evaluation strategy}: `steps',
     \textit{evaluation step}s: 500,
     \textit{save steps}: 1e6,
     \textit{gradient accumulation steps}: 16
\}. 

\subsection{Evaluation metrics without fine-tuning} \label{appendix: evaluation_without_finetuning}

\begin{table*}[h]
\centering
\small
{%
\begin{tabular}{lcccc}
\toprule
\textbf{Models}  & \textbf{ROUGE-1} & \textbf{ROUGE-2} & \textbf{ROUGE-L} & \textbf{SummaC}\\ \midrule
BART-base  & 0.212& 0.050& 0.107 & 0.315\\
BART-large-CNN   & 0.189& 0.028& 0.123 & 0.215\\
T5 large  & 0.247& 0.049& 0.135 & 0.212\\
BART-large-xsum-samsum  & 0.325& 0.117& 0.232 & 0.257\\
Pegasus-large & 0.240& 0.033& 0.133 & 0.219\\ \bottomrule
\end{tabular}%
}
\caption{ROUGE and SummaC values of the model generated summaries without fine-tuning.}
\label{tab: appendix_roUge_metric}
\end{table*}

\begin{table}[ht!]
    \centering
    \scriptsize
    \begin{tabular}{c>{\raggedright\arraybackslash}p{6cm}>{\raggedright\arraybackslash}p{6cm}}
    \toprule[0.5ex]
         Epochs(\#)&  Pegasus-large generated summary& BART-large-CNN generated summary\\ \midrule
         0 (No tuning)& Patient: Female Doctor: Please describe your social life at the college campus. Doctor: Describe your ability to attend to the task at hand or concentrate on daily tasks (academic, non-academic)? Doctor: Describe your ability to attend to the task at hand or concentrate on daily tasks (academic, non-academic)? But when I am in a bad mood I will distract myself from the task with social media and resume when I feel fine. Doctor: Have you noticed any difficulties with memory, such as unable to register new information, or not able to recall older personal/factual events?  & My mood is dynamic. It goes up and down for both valid and unknown reasons. I get upset on simple jokes and responses from my close circle. I feel supported very less even from family. And so I don't expect any support from anyone and try to figure out all alone. I'm reading books on self-development and self-improvement.\\ \midrule
         5&  Patient: Female Doctor: Please describe your social life at the college campus. Doctor: Describe your ability to attend to the task at hand or concentrate on daily tasks (academic, non-academic)? Doctor: Describe your ability to attend to the task at hand or concentrate on daily tasks (academic, non-academic)? But when I am in a bad mood I will distract myself from the task with social media and resume when I feel fine. Doctor: Have you noticed any difficulties with memory, such as unable to register new information, or not able to recall older personal/factual events?& Patient is a 19 year old girl. Her mood is generally happy but goes up and down throughout the day. She takes part in extracurricular activities and socializes with others. Her concentration and task attending ability is good. She feels difficulty with memory. For daily frustration she tries to connect with the Almighty. On a bad day, she sleeps for hours and feels better by singing a song or dancing. She does not feel supported by her family or friends. She is feeling stress, anxiety and depression symptoms. She reads self-development and self-improvement books.\\ \midrule
         10&  Patient: Female Doctor: Please describe your social life at the college campus. Doctor: Describe your ability to attend to the task at hand or concentrate on daily tasks (academic, non-academic)? Patient: I am mostly able to focus on my task and complete on time. But when I am in a bad mood I will distract myself from the task with social media and resume when I feel fine. Doctor: Have you noticed any difficulties with memory, such as unable to register new information, forgetting recent events, or not able to recall older personal/factual events?& Patient is a 19 year old girl. Her mood is generally happy but goes up and down throughout the day. She takes part in extracurricular activities and socializes with others. For daily frustration she tries to connect with the Almighty and sleeps. She feels academic pressure. Her concentration and task attending ability is okay but she feels difficulty with memory. She does not feel supported by her family and friends. On a bad day, she sleeps, cries and talks to herself. She is feeling stress, anxiety and depression symptoms. To overcome this she reads self-development and self-improvement books.\\ \midrule
         25&  Patient is a 19 year old girl. Her mood is generally happy but goes up and down throughout the day. She takes part in extracurricular activities and socializes with others. For daily frustration she tries to connect with the Almighty. She feels academic pressure. Her concentration and task attending ability is good. She does not feel difficulty with memory. She feels better by sleeping, singing songs, and dancing. She feels supported by her family and friends. On a bad day, she sleeps, talks to friends, and tries to push through. She is not feeling any stress, anxiety or depression symptoms.& Patient is a 19 year old girl. Her mood generally fluctuates throughout the day. She takes part in extracurricular activities and socializes with others. For daily frustration she does prayer and sleeps. She does not feel academic pressure. Her concentration and task attending ability is good. She feels difficulty with memory.She feels better by singing, dancing, sleeping, and feeling supported by her family and friends. On a bad day, she cries and talks to herself. She is feeling stress, anxiety, and depression symptoms. To overcome this she reads self-development and self-improvements books.\\ \midrule
         50&  Patient is a 19 year old girl. Her mood is generally happy but goes up and down throughout the day. She takes part in extracurricular activities and socializes with others. For daily frustration she prays and sleeps. She feels academic pressure. Her concentration and task attending ability is good. She feels difficulty with memory. She feels better by sleeping, singing, dancing, and talking to herself. She feels supported by her family and friends. On a bad day, she sleeps, talks to herself, and tries to push through. She is not feeling any stress, anxiety or depression symptoms.& Patient is a 19 year old girl. Her mood generally fluctuates throughout the day and goes up and down for unknown reasons. She takes part in extracurricular activities and socializes with others. For daily frustration she does prayer and sleeps. She feels academic pressure. Her concentration and task attending ability is normal. She feeling difficulty with memory. She feel better by singing, dancing and sleeping. She does not feel supported by her family and friends. On a bad day, she cries and talks to herself. She is feeling stress,anxiety, and depression symptoms. To overcome this she reads self-development and self-improvements books.\\ \midrule
         100&  Patient is a 19 year old girl. Her mood is generally happy but goes up and down throughout the day. She takes part in extracurricular activities and socializes with others. For daily frustration she does prayer and sleeps. She feels academic pressure. Her concentration and task attending ability is okay. She feels difficulty with memory. She feels better by singing, dancing and sleeping. She feels supported by her family and friends. On a bad day, she cries and talks to herself. She is feeling stress and anxiety symptoms such as difficulty sleeping and talking to herself. To overcome this she reads self-development and self-improvement books.& Patient is a 19 year old girl. Her mood is generally happy but goes up and down throughout the day. She takes part in extracurricular activities and socializes with others. For daily frustration she does prayer and sleeps. She feels academic pressure. Her concentration and task attending ability is normal. She feeling difficulty with memory.She feels better by singing, dancing and sleeping. She does not feel supported by her family and friends. On a bad day, she cries and talks to herself. She is feeling stress,anxiety, and depression symptoms. To overcome this she reads self-development and self-improvements books.\\ \bottomrule[0.5ex]
\end{tabular}
    \caption{Pegasus-large and BART-large-CNN generated summaries at different epochs on conversation given in Table \ref{tab:psychiatrist_patient_diaogue} in the Appendix}
\label{tab:model_varying_epochs}
\end{table}


\begin{table*}[]
    \centering
    \small
    \begin{subtable}{0.45\linewidth}
        \centering
        \begin{tabular}{lcc}
            \toprule
             & \textbf{Reviewer 1 }& \textbf{Reviewer 2 }\\ 
            \midrule
            \textbf{Reviewer 1} & 1.00& 0.24\\
            \textbf{Reviewer 2 }& 0.24& 1.00\\
            \bottomrule
        \end{tabular}
        \caption{On Pegasus model summaries}
        \label{tab:Cohen_clinical_pegasus}
    \end{subtable}
    \hfill
    \begin{subtable}{0.45\linewidth}
        \centering
        \begin{tabular}{lcc}
            \toprule
            & \textbf{Reviewer 1 }& \textbf{Reviewer 2 }\\ 
            \midrule
            \textbf{Reviewer 1} & 1.00& 0.19\\
            \textbf{Reviewer 2} & 0.19& 1.00\\
            \bottomrule
        \end{tabular}
        \caption{On BART-large-CNN model summaries}
        \label{tab:Cohen_non_clinical_bart}
    \end{subtable}
    \caption{Inter-rater reliability among clinical reviewers. Cohen's Kappa Coefficient on (a) Pegasus, (b) BART-large-CNN model generated summaries.}
    \label{tab:inter_rater_clinical}
\end{table*}

\begin{table*}[]
\centering
\setlength{\tabcolsep}{4pt}
\tiny
\begin{subtable}{0.48\linewidth}
\centering
\begin{tabular}{ccccccccccl} 
\toprule
 &  \textbf{A1}&  \textbf{A2}&  \textbf{A3}&  \textbf{A4}&  \textbf{A5}&  \textbf{A6}&  \textbf{A7}&  \textbf{A8}&  \textbf{A9}&\textbf{A10}\\ 
 \bottomrule
 \textbf{\textit{A1}}&  1.00&  0.43&  0.62&  0.43&  0.44&  0.58&  0.39&  0.46&  0.65&0.31 \\ 
 \textbf{\textit{A2}}&  0.43&  1.00&  0.38&  0.32&  0.41&  0.26&  0.25&  0.36&  0.27&0.35 \\ 
 \textbf{\textit{A3}}&  0.62&  0.38&  1.00&  0.35&  0.48&  0.66&  0.36&  0.57&  0.62&0.34 \\ 
 \textbf{\textit{A4}}&  0.43&  0.32&  0.35&  1.00&  0.32&  0.34&  0.45&  0.38&  0.35&0.30\\ 
 \textbf{\textit{A5}}&  0.44&  0.41&  0.48&  0.32&  1.00&  0.41&  0.45&  0.60&  0.41&0.53 \\ 
 \textbf{\textit{A6}}&  0.58&  0.26&  0.66&  0.34&  0.41&  1.00&  0.44&  0.70&  0.61&0.29 \\ 
 \textbf{\textit{A7}}&  0.39&  0.25&  0.36&  0.45&  0.45&  0.44&  1.00&  0.50&  0.32&0.34 \\ 
 \textbf{\textit{A8}}&  0.46&  0.36&  0.57&  0.38&  0.60&  0.70&  0.50&  1.00&  0.59&0.38 \\ 
 \textbf{\textit{A9}}&  0.65&  0.27&  0.62&  0.35&  0.41&  0.61&  0.32&  0.59&  1.00&0.26 \\ 
\textbf{\textit{A10}}& 0.31& 0.35& 0.34& 0.30& 0.53& 0.29& 0.34& 0.38& 0.26&1.00\\ \bottomrule
\end{tabular}
\caption{Pegasus Model}
\label{tab:Cohen_non_clinical_pegasus}
\end{subtable}
\hfill
\begin{subtable}{0.48\linewidth}
\centering
    \begin{tabular}{ccccccccccl} 
\toprule
 &  \textbf{A1}&  \textbf{A2}&  \textbf{A3}&  \textbf{A4}&  \textbf{A5}&  \textbf{A6}&  \textbf{A7}&  \textbf{A8}&  \textbf{A9}&\textbf{A10}\\ \bottomrule
 \textbf{\textit{A1}}&  1.00&  0.39&  0.78&  0.23&  0.52&  0.62&  0.55&  0.62&  0.50&0.49 \\ 
 \textbf{\textit{A2}}&  0.39&  1.00&  0.36&  0.28&  0.35&  0.44&  0.50&  0.47&  0.31&0.50\\ 
 \textbf{\textit{A3}}&  0.78&  0.36&  1.00&  0.32&  0.62&  0.57&  0.55&  0.72&  0.66&0.47 \\ 
 \textbf{\textit{A4}}&  0.23&  0.28&  0.32&  1.00&  0.37&  0.34&  0.37&  0.28&  0.29&0.30\\ 
 \textbf{\textit{A5}}&  0.52&  0.35&  0.62&  0.37&  1.00&  0.44&  0.46&  0.47&  0.39&0.52 \\ 
 \textbf{\textit{A6}}&  0.62&  0.44&  0.57&  0.34&  0.44&  1.00&  0.31&  0.51&  0.45&0.43 \\ 
 \textbf{\textit{A7}}&  0.55&  0.50&  0.55&  0.37&  0.46&  0.31&  1.00&  0.49&  0.38&0.40\\ 
 \textbf{\textit{A8}}&  0.62&  0.47&  0.72&  0.28&  0.47&  0.51&  0.49&  1.00&  0.54&0.41 \\ 
 \textbf{\textit{A9}}&  0.50&  0.31&  0.66&  0.29&  0.39&  0.45&  0.38&  0.54&  1.00&0.36 \\ 
\textbf{\textit{A10}}& 0.49& 0.50& 0.47& 0.30& 0.52& 0.43& 0.40& 0.41& 0.36&1.00\\ \bottomrule
\end{tabular}
\caption{BART-large-CNN Model}
\label{tab:Cohen_non_clinical_bart}
\end{subtable}
\caption{Inter-rater Reliability (non-Clinical Annotators) - Cohen's Kappa Coefficient on (a) Pegasus Model and (b) BART-large-CNN Model}
\label{tab:Cohen_non_clinical}
\end{table*}

\subsection{Summary evaluation}

\begin{figure}[H]
    \centering
    \begin{subfigure}{0.45\textwidth}
        \includegraphics [width=\textwidth]{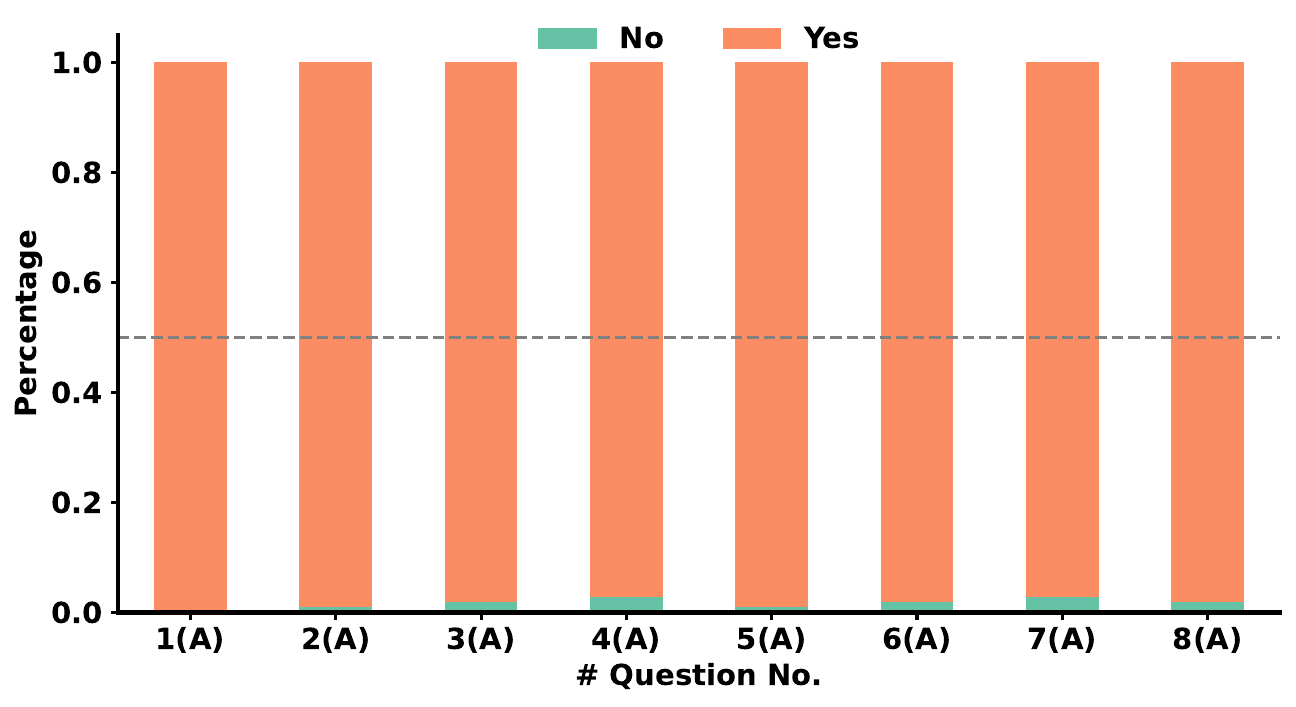}
        \caption{On Pegasus summaries}
        \label{fig: HE_Pegasus_A}
    \end{subfigure} 
        \begin{subfigure}{0.45\textwidth}
        \includegraphics [width=\textwidth]{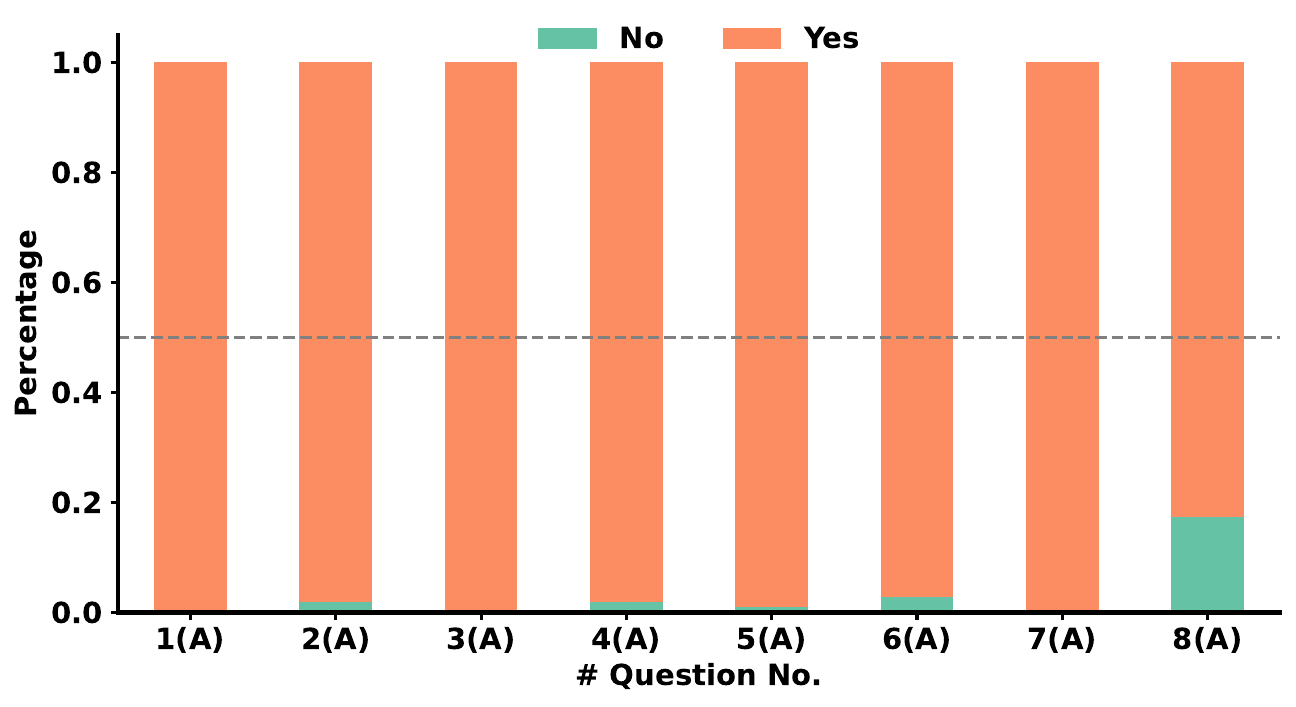}
        \caption{On BART-large-CNN summaries}
        \label{fig: HE_BART-LARGE_A}
    \end{subfigure} 
    
    \begin{subfigure}{0.45\textwidth}
        \includegraphics [width=\textwidth]{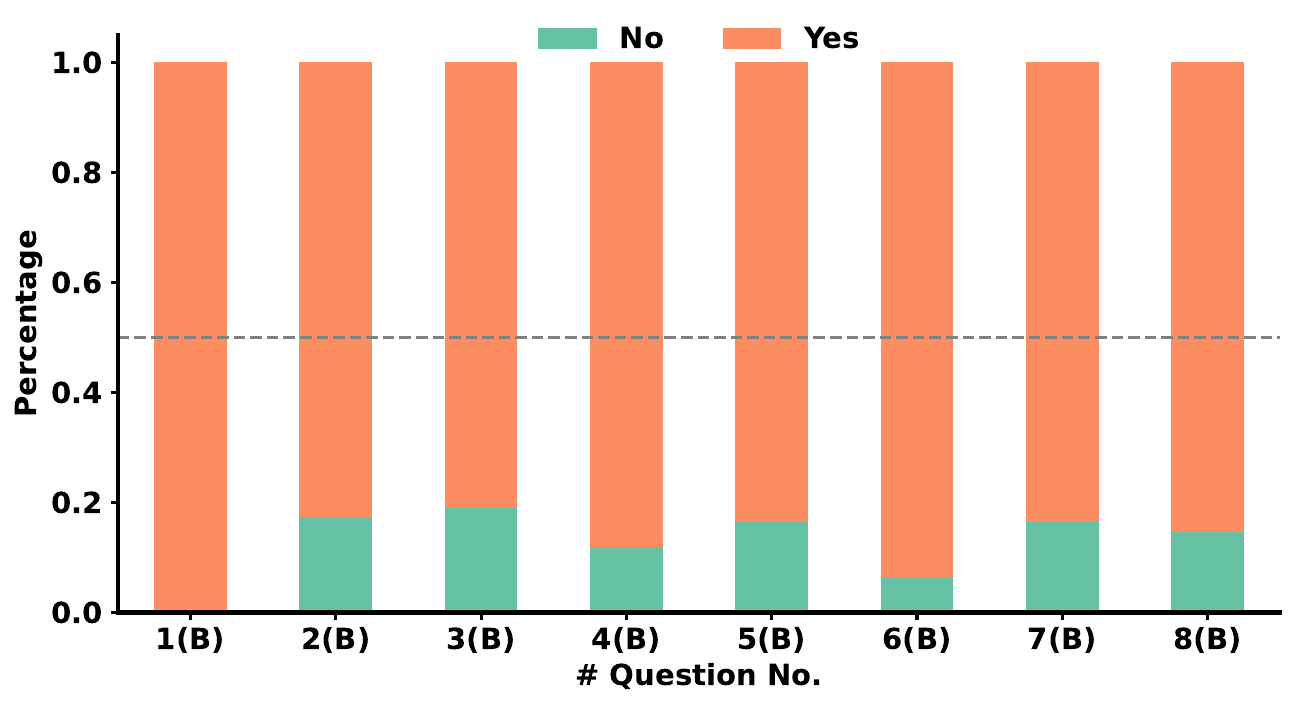}
        \caption{On Pegasus summaries}
        \label{fig: HE_Pegasus_B}
    \end{subfigure} 
        \begin{subfigure}{0.45\textwidth}
        \includegraphics [width=\textwidth]{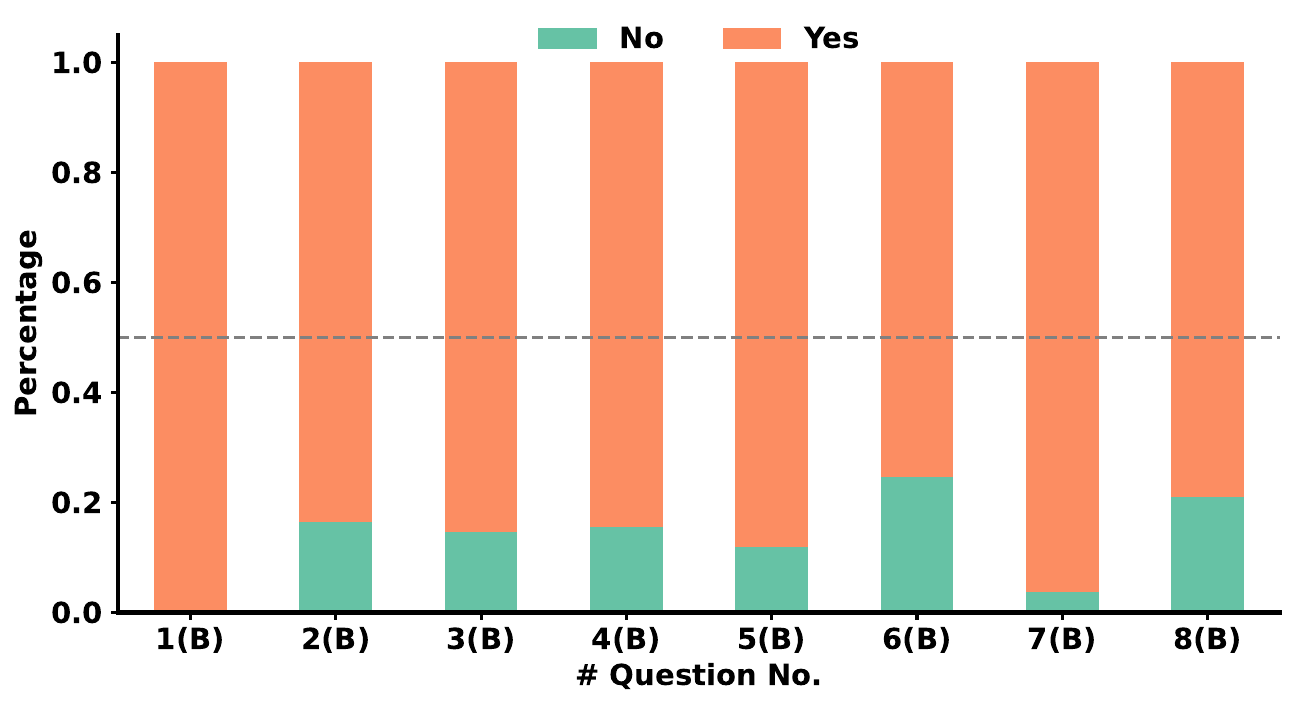}
        \caption{On BART-large-CNN summaries}
        \label{fig: HE_BART-LARGE_B}
    \end{subfigure} 
    \caption{
Fine-grained \textbf{human evaluation} of Pegasus-large and BART-large-CNN summaries. (a) and (b) show the percentage of summaries capturing the following parameters of the input conversation: 1(A) gender, 2(A) mood, 3(A) social life, 4(A) academic pressure, 5(A) concentration ability, 6(A) difficulty with memory, 7(A) strategies to feel better, and 8(A) mental disorders with Pegasus-large and BART-large-CNN, respectively.
Similarly, (c) and (d) show the percentage of summaries consistent with the input conversation on the following parameters: 1(B) gender, 2(B) mood, 3(B) social life, 4(B) academic pressure, 5(B) concentration ability, 6(B) difficulty with memory, 7(B) strategies to feel better, and 8(B) mental disorders with Pegasus model, and BART-large-CNN model, respectively. 
    }
    \label{fig:HE_question_wise}
\end{figure}

\begin{figure}[H]
    \centering
    \begin{subfigure}{0.45\textwidth}
        \includegraphics [width=\textwidth]{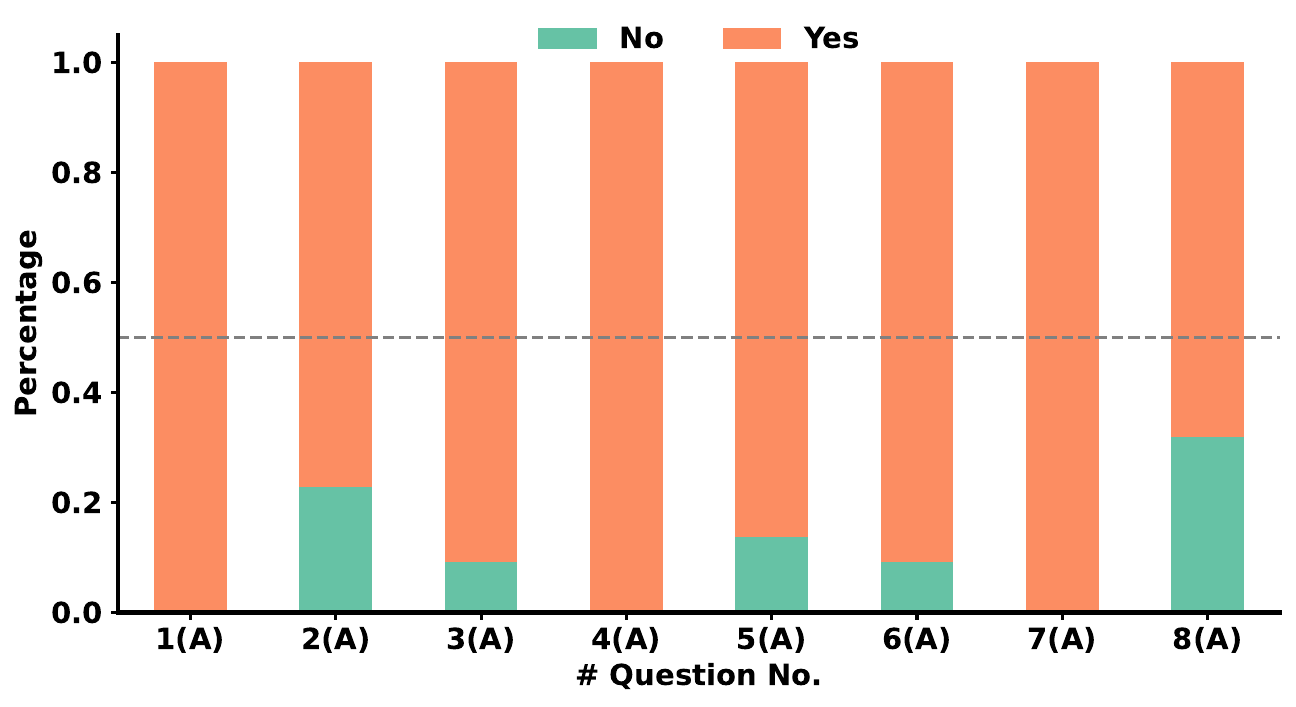}
             \caption{On Pegasus summaries}
        \label{}
    \end{subfigure} 
        \begin{subfigure}{0.45\textwidth}
        \includegraphics [width=\textwidth]{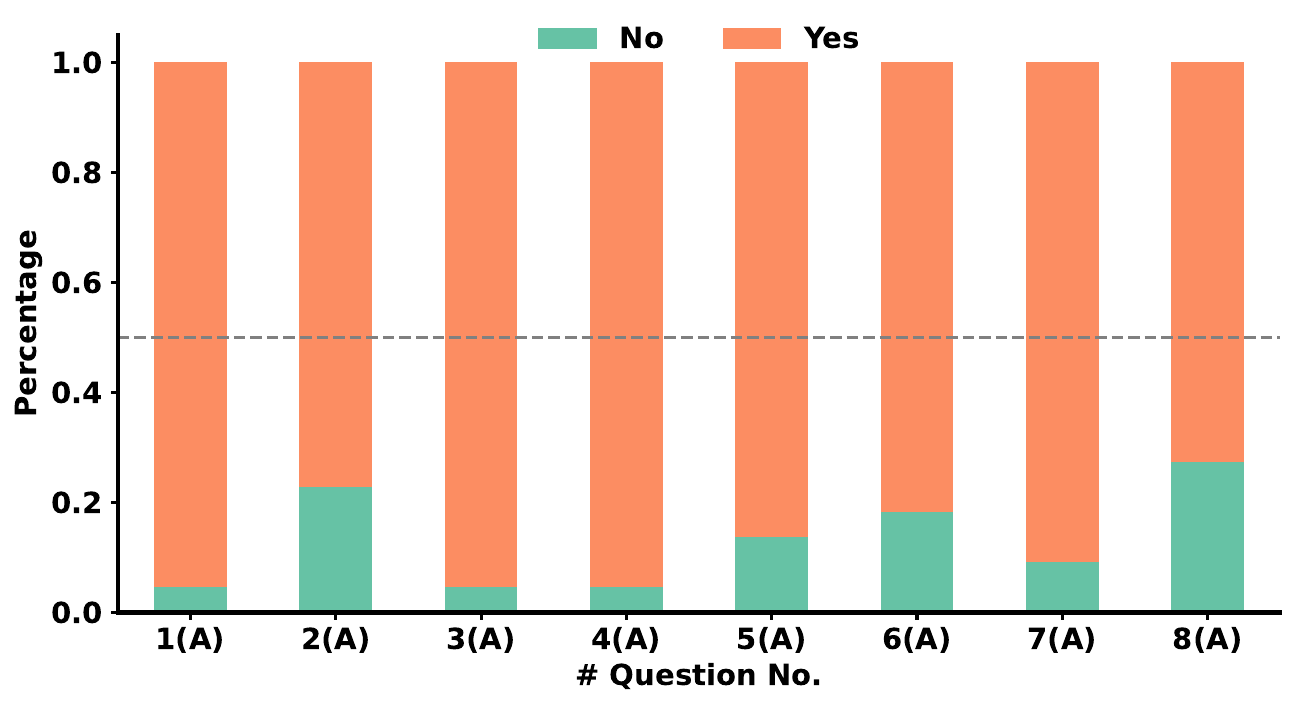}
        \caption{On BART-large-CNN summaries}
        \label{}
    \end{subfigure} 

    \begin{subfigure}{0.45\textwidth}
        \includegraphics [width=\textwidth]{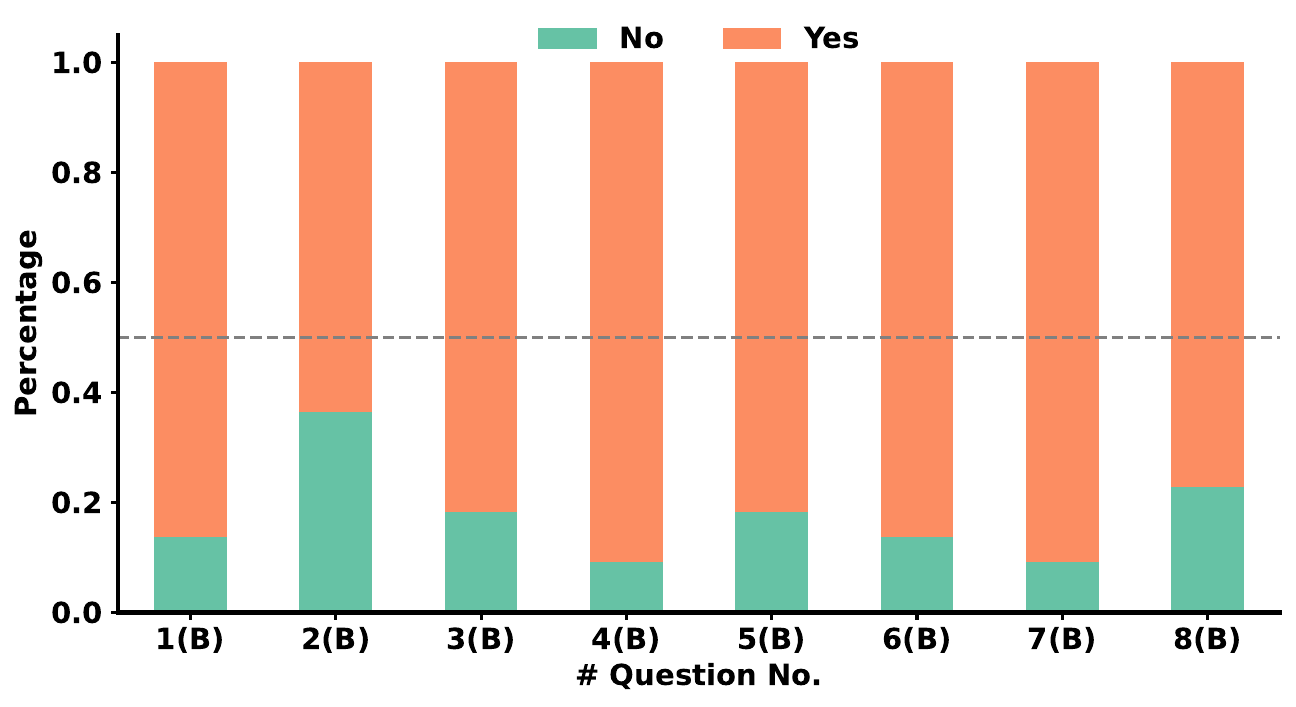}
        \caption{On Pegasus summaries}
        \label{}
    \end{subfigure} 
        \begin{subfigure}{0.45\textwidth}
        \includegraphics [width=\textwidth]{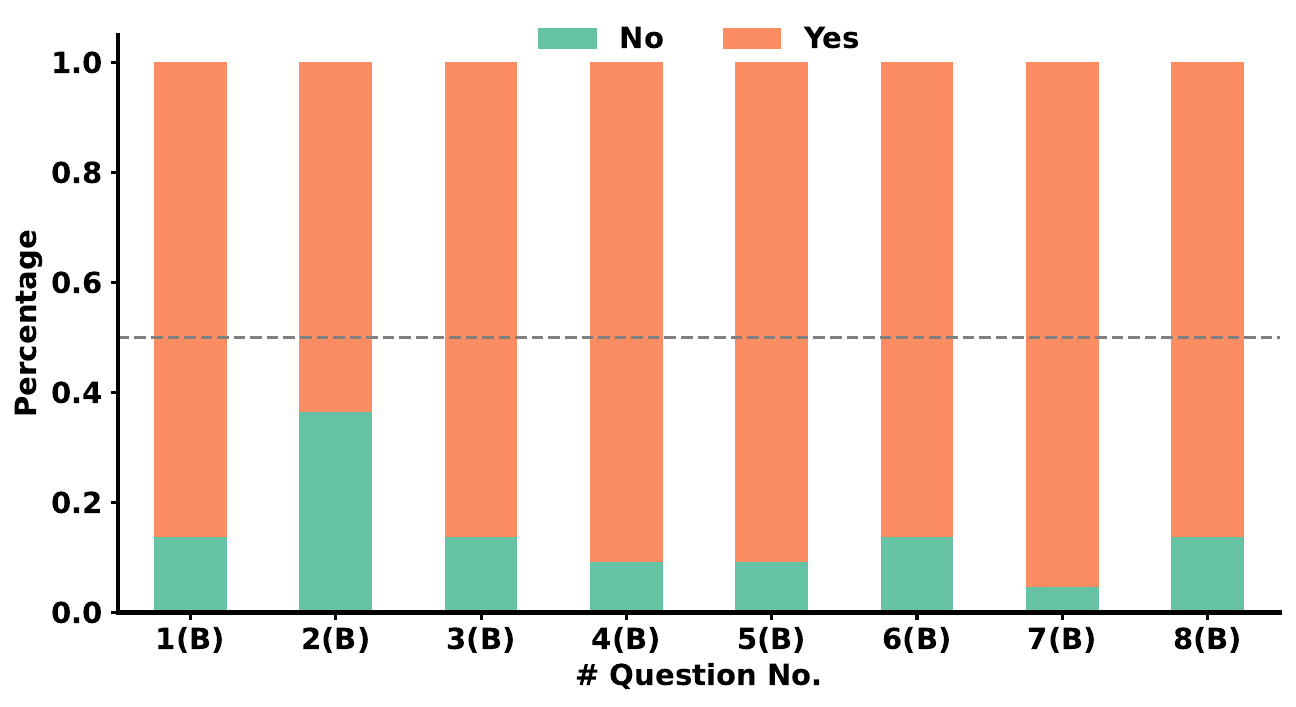}
        \caption{On BART-large-CNN summaries}
        \label{}
    \end{subfigure} 
    \caption{
    Fine-grained \textbf{LLM evaluation} of Pegasus-large and BART-large-CNN summaries. (a) and (b) show the percentage of summaries capturing the following parameters of the input conversation: 1(A) gender, 2(A) mood, 3(A) social life, 4(A) academic pressure, 5(A) concentration ability, 6(A) difficulty with memory, 7(A) strategies to feel better, and 8(A) mental disorders with Pegasus-large and BART-large-CNN, respectively.
Similarly, (c) and (d) show the percentage of summaries consistent with the input conversation on the following parameters: 1(B) gender, 2(B) mood, 3(B) social life, 4(B) academic pressure, 5(B) concentration ability, 6(B) difficulty with memory, 7(B) strategies to feel better, and 8(B) mental disorders with Pegasus model, and BART-large-CNN model, respectively. 
    }
\label{fig: llm_evaluation_plot}
\end{figure}

\subsection{Prompt} \label{appendix: prompt}
  
\begin{small}
\begin{spverbatim}
Consider yourself as an individual who is proficient in English. You need to rate two summaries generated for the given conversation data on four parameters listed below:
1.Fluency: Is the summary well structured, free from awkward phrases, and grammatically correct?
2.Completeness: Does the summary cover all relevant aspects of the conversation?
Metric
1	2	3	4	5
Fluency	Not fluent at all	Slightly fluent	Moderately fluent	Quite fluent	Very fluent
Completeness	Not complete at all	Slightly complete	Moderately complete	Quite complete	Very complete

3.Hallucinations: Does the summary contain any extra information that a user did not present? Simply put, this metric captures to what extent the generated summary contains new information that is not a part of the user conversation. For example, if a user does not mention anything about friends during the conversation, and the summary mentions something related to friends, then it is an example of hallucination.
4.Contradiction: Does the summary contradict the information provided by a user? Simply put, this metric captures to what extent the summary contradicts the user conversation. For example, if a user says that he has a good memory and the summary says that the participant has a poor memory, it is an example of contradiction.
Metric
1	2	3	4	5
Hallucination	No hallucination	Mild hallucination	Moderate hallucination	Severe hallucination	Extremely severe hallucination
Contradiction	No Contradiction	Mild Contradiction	Moderate Contradiction	Severe Contradiction	Extremely severe Contradiction.
Please stick with the rating, dont provide any reasoning. Also, You need to answer in Yes or No for the following questions for both the summary:-
1. Gender
1(a)Does the summary capture the gender of the user?
1(b)Is the summary data consistent with the provided conversation? 
2. Mood
2(a)Does the summary capture the mood of the user? 
2(b)Is the summary data consistent with the provided conversation?
3. Social Life
3(a)Does the summary capture the social life of the user?
3(b)Is the summary data consistent with the provided conversation?
4. Academic Pressure
4(a)Does the summary capture the academic pressure of the user?
4(b)Is the summary data consistent with the provided conversation?
5. Concentration ability
5(a)Does the summary capture the concentration ability of the user?
5(b)Is the summary data consistent with the provided conversation?
6. Difficulty with memory
6(a)Does the summary capture the memory difficulty of the user?
6(b)Is the summary data consistent with the provided conversation?
7. Strategies to feel better
7(a)Does the summary capture the strategies employed by the user to feel better?
7(b)Is the summary data consistent with the provided conversation?
8. Mental Disorder
8(a)Does the summary capture the symptoms of mental disorders stated by the user?
8(b)Is the summary data consistent with the provided conversation?

The results should look like this
---------Evaluation 1------------------
#	Completeness	Fluency		Hallucination	Contradiction
Summary1	<Completeness_point>	<Fluency_point>		<Hallucination_point>	<Contradiction_point>
Summary2	<Completeness_point>	<Fluency_point>		<Hallucination_point>	<Contradiction_point>
---------Evaluation 2------------------
Parameters	Summary1	Summary2
1(a)	<Summary1(Yes/No)>	<Summary2(Yes/No)>
1(b)	<Summary1(Yes/No)>	<Summary2(Yes/No)>
2(a)	<Summary1(Yes/No)>	<Summary2(Yes/No)>
2(b)	<Summary1(Yes/No)>	<Summary2(Yes/No)>
3(a)	<Summary1(Yes/No)>	<Summary2(Yes/No)>
3(b)	<Summary1(Yes/No)>	<Summary2(Yes/No)>
4(a)	<Summary1(Yes/No)>	<Summary2(Yes/No)>
4(b)	<Summary1(Yes/No)>	<Summary2(Yes/No)>
5(a)	<Summary1(Yes/No)>	<Summary2(Yes/No)>
5(b)	<Summary1(Yes/No)>	<Summary2(Yes/No)>
6(a)	<Summary1(Yes/No)>	<Summary2(Yes/No)>
6(b)	<Summary1(Yes/No)>	<Summary2(Yes/No)>
7(a)	<Summary1(Yes/No)>	<Summary2(Yes/No)>
7(b)	<Summary1(Yes/No)>	<Summary2(Yes/No)>
8(a)	<Summary1(Yes/No)>	<Summary2(Yes/No)>
8(b)	<Summary1(Yes/No)>	<Summary2(Yes/No)>
9(a)	<Summary1(Yes/No)>	<Summary2(Yes/No)>
9(b)	<Summary1(Yes/No)>	<Summary2(Yes/No)>
\end{spverbatim}
\end{small}

\begin{figure*}[H]
    \centering
    \begin{subfigure}{0.45\textwidth}
        \includegraphics [width=\textwidth]{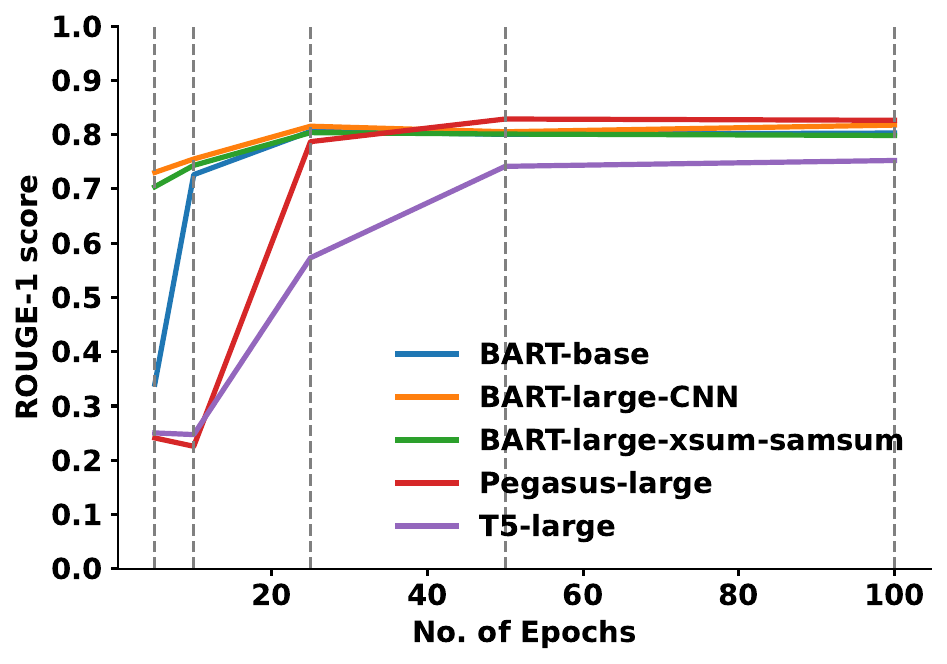}
        \caption{ROUGE-1 score}
        \label{}
    \end{subfigure} 
    \begin{subfigure}{0.45\textwidth}
        \includegraphics [width=\textwidth]{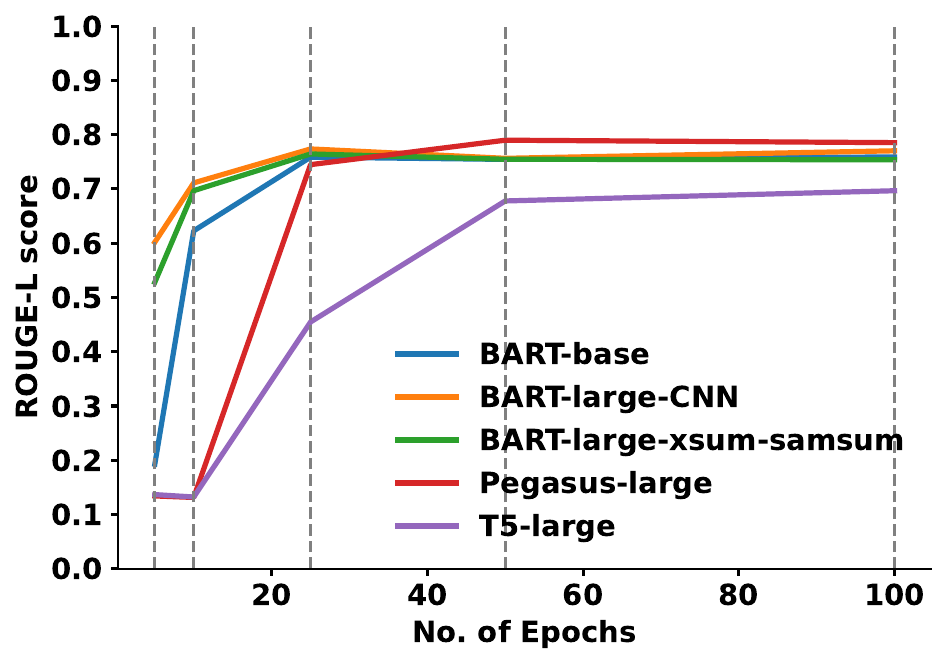}
        \caption{ROUGE-L score}
        \label{}
    \end{subfigure} 
    \caption{ROUGE-1 and ROUGE-L obtained after fine-tuning on BART-base, BART-large-CNN, T5 large, BART-large-xsum-samsum, and Pegasus-large with epochs = [5,10,25,50,100]}
    \label{fig:varying epochs}
\end{figure*}

    

    


\begin{table}
\tiny
\subcaptionbox{Conversation \label{}}{
\begin{tabular}{p{8cm}}
\toprule[0.5ex]
\textcolor{red}{Doctor}: What is your patient ID? \\
\textcolor{green}{Patient}: 1001 \\
\textcolor{red}{Doctor}: What is your age? \\
\textcolor{green}{Patient}: 32 \\
\textcolor{red}{Doctor}: What is your gender? \\
\textcolor{green}{Patient}: Female \\
\textcolor{green}{Patient}: "Okay" \\
\textcolor{red}{Doctor}: "Hello" \\
\textcolor{red}{Doctor}: "What are your main problems recently?" \\
\textcolor{green}{Patient}: "I haven't been feeling well recently, and I feel a little tight in my chest" \\
\textcolor{red}{Doctor}: "Have you ever gone to the hospital to see a doctor?" \\
\textcolor{green}{Patient}: "Not yet, I don't have much time recently" \\
\textcolor{green}{Patient}: "Maybe it will take two weeks to go" \\
\textcolor{red}{Doctor}: "Hmm, let's take some time to see if you have any emotional problems recently" \\
\textcolor{green}{Patient}: "There's nothing wrong with my mood, I just feel mentally tired recently" \\
\textcolor{red}{Doctor}: "Do you feel tired without doing anything?" \\
\textcolor{green}{Patient}: "I feel like this, I don't want to move" \\
\textcolor{red}{Doctor}: "Then do you feel like you don't want to work?" \\
\textcolor{green}{Patient}: "I don't have enough energy to work" \\
\textcolor{green}{Patient}: "Yes" \\
\textcolor{green}{Patient}: "But I have to work" \\
    \textcolor{green}{Patient}: "It's quite stressful" \\
\textcolor{red}{Doctor}: "Have you ever felt that you have lost interest in your past hobbies?" \\
\textcolor{green}{Patient}: "I should still be interested" \\
\textcolor{green}{Patient}: "I just don't have enough time to develop my hobbies" \\
\textcolor{red}{Doctor}: "Sleep, eat, etc." \\
\textcolor{red}{Doctor}: "Is everything normal?" \\
\textcolor{green}{Patient}: "fairly normal" \\
\textcolor{red}{Doctor}: "Will you feel dizzy or nauseous?" \\
\textcolor{green}{Patient}: "I get dizzy occasionally" \\
\textcolor{red}{Doctor}: "Do you feel lack of confidence? You are always worried about not doing well" \\
\textcolor{green}{Patient}: "No, I don't have time to worry about this or that" \\
\textcolor{red}{Doctor}: "It sounds like you are doing well lately" \\
\textcolor{red}{Doctor}: "The consultation ends here" \\ 
\bottomrule[0.5ex]
\end{tabular}
}
\subcaptionbox{Pegasus-large generated summary \label{}}{
\begin{tabular}{>{\raggedright\arraybackslash}p{6cm}}
\toprule[0.5ex]
Female Patient: "Okay" Doctor: "Hello" Doctor: "What are your main problems recently?" Patient: "I haven't been feeling well recently, and I feel a little tight in my chest" Doctor: "Have you ever gone to the hospital to see a doctor?" Patient: "Not yet, I don't have much time recently" Patient: "Hmm it will take two weeks to go" Doctor: "Maybe, let's take some time to see if you have any emotional problems recently" Patient: "There's nothing wrong with my mood, I just feel mentally tired recently" Doctor: "Do you feel tired without doing anything?" Patient: "I feel like this, I don't want to move" Doctor: "Then do you feel like you don't want to work?" Patient: "I don't have enough energy to work" Patient: "Yes" Patient: "But I have to work" Patient: "It's quite stressful" Doctor: "Have you ever felt that you have lost interest in your past hobbies?" Patient: "I should still be interested" Patient: "I just don't have enough time to develop my hobbies" Doctor: "Sleep, eat, etc."\\
\bottomrule[0.5ex]
\end{tabular}
}
\subcaptionbox{BART-large-CNN generated summary \label{}}{
\begin{tabular}{>{\raggedright\arraybackslash}p{6cm}}
\toprule[0.5ex]
 Patient is a 32 year old girl. She is feeling a little tight in her chest and haven't been feeling well recently. She hasn't gone to the hospital to see a doctor. Her mood is good, she just feel mentally tired. She doesn't want to work because she doesn't have energy and energy isn't good. She feels dizzy and nauseous. She does not feel lack of confidence. She thinks she is doing well.\\
\bottomrule[0.5ex]
\end{tabular}
}
\caption{Finetuned Pegasus-large and BART-large-CNN generated summary on a sample Chinese psychological conversation taken from \citep{yao2022d4}}
\label{tab:chinese_dialogue_1}
\end{table}

\begin{table*}[t]
\small
\centering
{%
\begin{tabular}{llcccc}
\toprule
  && \textbf{Completeness} & \textbf{Fluency} & \textbf{Hallucination} & \textbf{Contradiction} \\ 
  && \textbf{($\mu$, $\sigma$)} & \textbf{($\mu$, $\sigma$)} & \textbf{($\mu$, $\sigma$)} & \textbf{($\mu$, $\sigma$)} \\ \midrule
 \multirow{2}{*}{\textbf{D4}}&\textit{Pegasus-large}& (2.82, 1.40)& (2.96, 1.55)& (1.86 1.37)& (2.66, 1.67)\\
 & \textit{BART-large-CNN}& (4.46, 0.64)& (4.62, 0.53)& (1.60, 0.78)& (1.66, 0.74)\\ \midrule
 \multirow{2}{*}{\textbf{ESC}}& \textit{Pegasus-large}& (2.76, 1.17)& (3.06, 1.20)& (1.68, 1.07)& (1.92, 1.08)\\
 & \textit{Bart-large-CNN}& (4.14, 0.98)& (4.60, 0.60)& (1.62, 1.06)& (1.80, 1.08)\\ \bottomrule
\end{tabular}%
}
\caption{Average non-clinician human evaluation scores on D4 and ESC datasets with Pegasus-large and BART-large-CNN. For \textit{Completeness} and \textit{Fluency}, a rating closer to 5 indicates the best, whereas for \textit{Hallucination} and \textit{Contradiction}, a rating closer to 1 is preferable.}
\label{tab:generalizability_evaluator_Score}
\end{table*}


\begin{table}[H]
\tiny
\subcaptionbox{Conversation \label{}}{
\begin{tabular}{p{8cm}}
\toprule[0.5ex]
\textcolor{red}{Doctor}: What is your Patient ID? \\
\textcolor{green}{Patient}: 1004 \\
\textcolor{red}{Doctor}: What is your age? \\
\textcolor{green}{Patient}: 18 \\
\textcolor{red}{Doctor}: What is your gender? \\
\textcolor{green}{Patient}: Male \\
\textcolor{green}{Patient}: Hello \\
\textcolor{red}{Doctor}: Hi there, how can I help you?\\
\textcolor{green}{Patient}: I would like some help with the problem I am facing.\\
\textcolor{red}{Doctor}: OK, sure. Can you tell me what the problem is? I'll do my best to help.\\
\textcolor{green}{Patient}: Well, I am going into my next college semester next month, and I am very frightened about a calculus class I have to take. It's an honors course and I am very worried that I will not do well.\\
\textcolor{red}{Doctor}: I can understand that. It must be an important exam for you. Do you enjoy calculus?\\
\textcolor{red}{Doctor}: I'm terrible at anything with numbers, myself!\\
\textcolor{green}{Patient}: I don't remember, I took an easy calculus course in high school but that was a couple years ago. I only got a B there, so I'm worried about taking an honors one. I have to take it for my degree goal.\\
\textcolor{red}{Doctor}: A B is a great result!\\
\textcolor{red}{Doctor}: Are there any revision courses that you can access to help refresh your skills?\\
\textcolor{green}{Patient}: I don't think so, the campus is closed due to the pandemic. There are a few online tutors but they are almost always busy.\\
\textcolor{red}{Doctor}: This pandemic has really made it difficult to get anything done hasn't it.\\
\textcolor{red}{Doctor}: Do you know of anyone else who is looking to take the course that you could revise together with?\\
\textcolor{red}{Doctor}: I always find it easier to work through things like this with someone else.\\
\textcolor{green}{Patient}: Yes, the pandemic has affected many of my friends and their families, very difficult these times are.\\
\textcolor{green}{Patient}: No, none of my friends are taking the class/ none of them have ever taken calculus.\\
\textcolor{red}{Doctor}: There seem to be some really good online courses out there. It might be worth sending out inquiries to a number of tutors, asking them to get in touch as soon as they are free, and in the meantime working through something online to help build up your confidence?\\
\textcolor{green}{Patient}: That sounds like some good advice to me.\\
\textcolor{green}{Patient}: Maybe I'll try working on it for my confidence like you said.\\
\textcolor{red}{Doctor}: I bet once you get into it again it will all come back to you. Sometimes getting started is the hardest part.\\
\textcolor{green}{Patient}: I think you're right!\\
\textcolor{green}{Patient}: Do you believe in me? That I will pass well?\\
\textcolor{red}{Doctor}: I think that if you got a B once, you can do it again, and if you really want it, which you clearly do, you can do anything! Aim for an A, you can do it!\\
\textcolor{green}{Patient}: Thank you very much you've been very helpful.\\
\textcolor{green}{Patient}: God bless you, have a good day.\\
\textcolor{red}{Doctor}: And you too, good luck!\\
\bottomrule[0.5ex]
\end{tabular}
}
\subcaptionbox{Pegasus-large Generated Summary \label{}}{
\begin{tabular}{>{\raggedright\arraybackslash}p{6cm}}
\toprule[0.5ex]
Patient is an 18 year old boy. His age is not known but he does not take calculus. He is worried about taking an honors calculus course. He does not know anyone who takes calculus. He does not take revision courses. He does not know anyone who takes calculus. He is worried about taking an honors calculus course. He does not know anyone who takes calculus. He is worried about taking an honors calculus course. He does not know anyone who takes calculus. He does not know anyone who takes calculus. He is worried about taking an honors calculus course. He does not know anyone who takes calculus. He does not know anyone who takes calculus. He is worried about taking an honors calculus course. He does not know anyone who takes calculus. He does not know anyone who takes calculus. He is worried about taking an honors calculus course. He does not know anyone who takes calculus. He does not know anyone who takes calculus. He is worried about taking an honors calculus course. He does not know anyone who takes calculus. He does not know anyone who takes calculus. He does not know anyone who takes calculus. He is worried about taking an honors calculus course. He does not know anyone who takes calculus. He does not know anyone who takes calculus. He is worried about taking an honors calculus course. He does not know anyone who takes calculus. He does not know anyone who takes calculus.\\
\bottomrule[0.5ex]
\end{tabular}
}
\subcaptionbox{BART-large-CNN generated summary \label{}}{
\begin{tabular}{>{\raggedright\arraybackslash}p{6cm}}
\toprule[0.5ex]
Patient is an 18 year old boy. He is worried about a calculus exam for an honors course. He does not enjoy calculus and does not think he will do well. There are no revision courses available as the campus is closed due to the pandemic and there are a few online tutors. He knows no one who is taking the class and none of his friends have ever taken calculus. He feels alone. He takes the exam and gets a B. He hopes to get an A. He asks the Doctor for help with his confidence. The Doctor gives him some advice and he hopes he will pass.\\
\bottomrule[0.5ex]
\end{tabular}
}
\caption{Finetuned Pegasus-large and BART-large-CNN generated summary on an Empathy Support Conversation (ESC) conversation taken from \citep{liu2021towards}}
\label{tab:ESC}
\end{table}

\begin{table*}[t]
\centering
\scriptsize
\begin{tabular}{ccccc} 
\toprule
\textbf{System Configuration}&  \textbf{Model}&  \textbf{RAM usage before}&  \textbf{RAM usage while running}& \textbf{Response time}\\ 
&  &  (GB)&  (GB)& (s)\\ \midrule

\multirow{2}{*}{Processor - i5-1135G7 @ 2.40GHz, RAM - 16GB}&  \textit{Pegasus-large}&  6.65&  8.57& 32.63\\
& \textit{BART-large-CNN}& 6.75& 8.23&22.03\\ \midrule

\multirow{2}{*}{Processor - i7-10700 @ 2.90GHz, RAM - 16GB}& \textit{Pegasus-large} & 14.04& 14.75&30.02\\ 
& \textit{BART-large-CNN}& 13.21& 14.99&22.74\\ \midrule
\multirow{2}{*}{Processor - i9-12900K  @ 3.20GHz, RAM - 64GB}& \textit{Pegasus-large} & 27.08& 29.29&16.44\\
&  \textit{BART-large-CNN}&  25.39&  28.12& 10.59\\ \bottomrule
\end{tabular}
\caption{
Response time and random Access Memory(RAM) consumption before and during execution of  models (Pegasus-large, BART-large-CNN) on three different systems with varying configuration.
}
\label{tab:low_end}
\end{table*}

\section{Discussion} \label{appendix: discussion}

This appendix section sheds insights and intuitions we gained during our study.  

\subsection{Comparison with the previous work}  
Our work represents the first attempt to summarize psychological conversation data, which differs from traditional text summarization. However, it shares similarities with dialogue summarization, such as summarizing conversations between individuals or medical dialogues between doctors and patients. Table \ref{tab:comparison_with_previous_work} illustrates the positioning of our work in the landscape of text summarization within healthcare. To the best of our knowledge, we only identified the work by Yao et al. \citep{yao2022d4}, where they summarized symptoms using psychological conversation data. Furthermore, our fine-tuned model consistently generated fluent and comprehensive summaries, even when applied to datasets utilized by Yao et al.

It is important to acknowledge that the studies presented in Table \ref{tab:comparison_with_previous_work} utilized different datasets. In contrast, we demonstrated the effectiveness of our model on both our dataset and publicly available psychological conversational datasets, D4 and ESC. However, it is important to note that existing studies have their own specific objectives beyond solely summarizing entire conversations. While our work primarily aims at generating summaries of psychological conversations, it encounters its own challenges, such as dealing with lengthy conversation data, resulting in longer utterances. This distinction is essential to consider when evaluating the performance and applicability of our model compared to previous studies.

\begin{table}[ht]
    \centering
    \scriptsize
    \begin{tabular}{lllccc}
    \toprule
         \textbf{Reference}&  \textbf{Model}&  \textbf{Dataset}&  \textbf{ROUGE-1}& \textbf{ROUGE-2} &\textbf{ROUGE-L}\\ 
         &  (own/ fine-tuned)&  &  &  &\\ \midrule
         \citep{krishna2021generating}&  fine-tuned&  Medical (Own prepared)&  0.57&  0.29&0.38\\
         &  fine-tuned&  AMI medical corpus&  0.45&  0.17&0.24\\ \midrule
         \citep{michalopoulos2022medicalsum}&  own&  MEDIQA 2021 - history of present illness&  0.48&  -&0.35\\
         &  own&  MEDIQA 2021 - physical examination
&  0.68&  -&0.64\\
         &  own&  MEDIQA 2021 - assessment and plan
&  0.44&  -&0.37\\
         &  own&  MEDIQA 2021 - diagnostic imaging results
&  0.27&  -&0.26\\ \midrule
 \citep{song2020summarizing}& fine-tuned& Medical problem Description& 0.91& 0.87&0.91\\
 & fine-tuned& Medical diagnosis or treatment & 0.80& 0.72&0.80\\
 & fine-tuned& Medical problem Description
& 0.91& 0.87&0.91\\
 & fine-tuned& Medical diagnosis or treatment& 0.81& 0.73&0.81\\ \midrule
\citep{zhang2021leveraging} & fine-tuned& Doctor patient conversation & 0.46& 0.19&0.44\\ \midrule
\citep{yao2022d4} & fine-tuned& Chinese psychological conversation & -& -&0.26\\ \midrule[0.25ex]
\multirow{2}{*}{\textbf{Our Work}} & \textbf{Pegasus-large}& \multirow{2}{*}{\textbf{Psychological conversation (own)}} & \textbf{0.83}& \textbf{0.71}& \textbf{0.79}\\ 
 & \textbf{BART-large-CNN}& & \textbf{0.81}& \textbf{0.69}&\textbf{0.77}\\\bottomrule
    \end{tabular}
\caption{Comparison of our best model results in terms of ROUGE with existing works.}
\label{tab:comparison_with_previous_work}
\end{table}

\subsection{Fine-tuned Pegasus-large versus fine-tuned BART-large-CNN models performance} 

\begin{figure}[h!]
    \centering
    \begin{subfigure}{0.95\textwidth}
        \includegraphics [width=\textwidth]{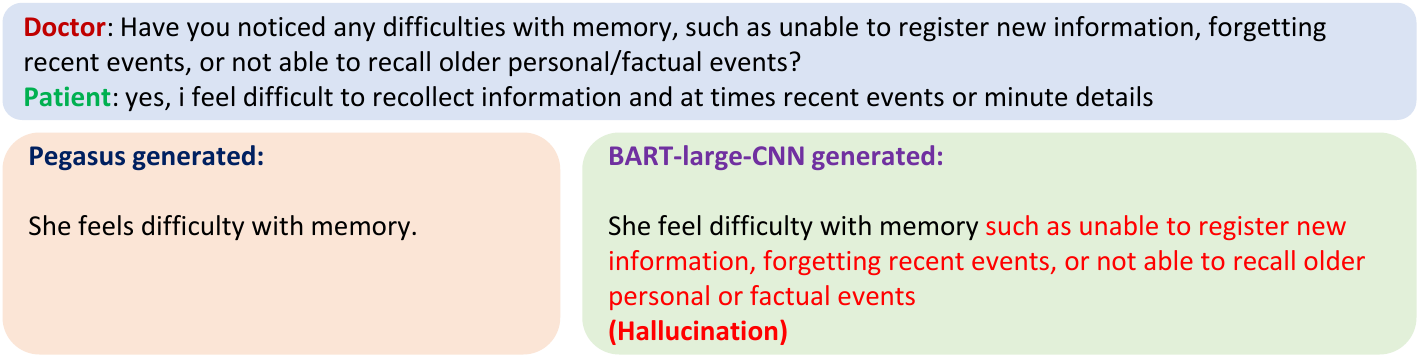}
        \caption{}
        \label{fig: excerpt_1}
    \end{subfigure} 
    
        \begin{subfigure}{0.95\textwidth}
        \includegraphics [width=\textwidth]{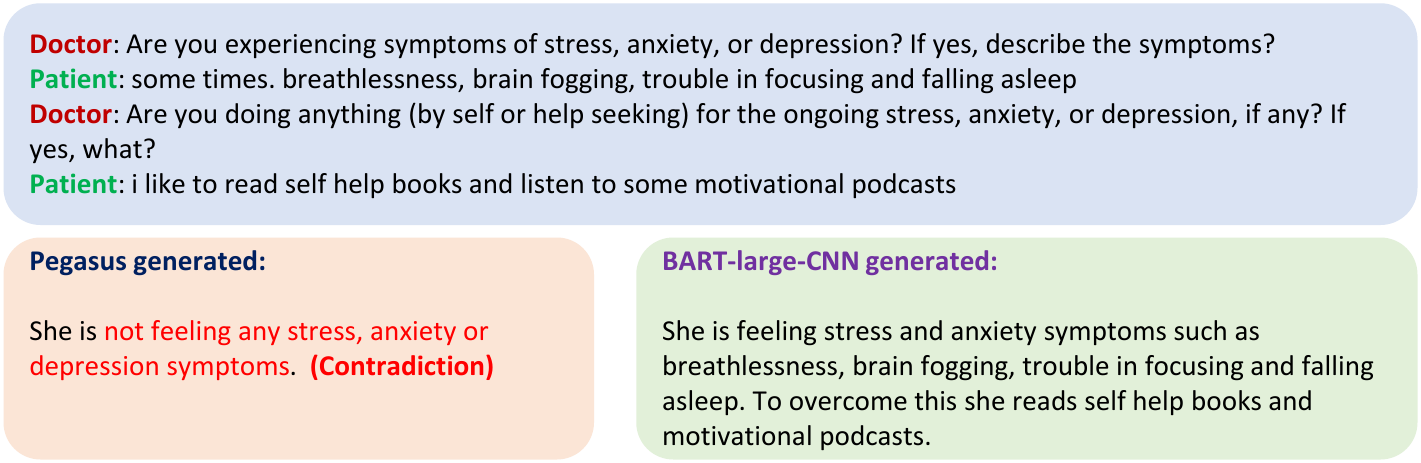}
        \caption{}
        \label{fig: excerpt_2}
    \end{subfigure}

        \begin{subfigure}{0.95\textwidth}
        \includegraphics [width=\textwidth]{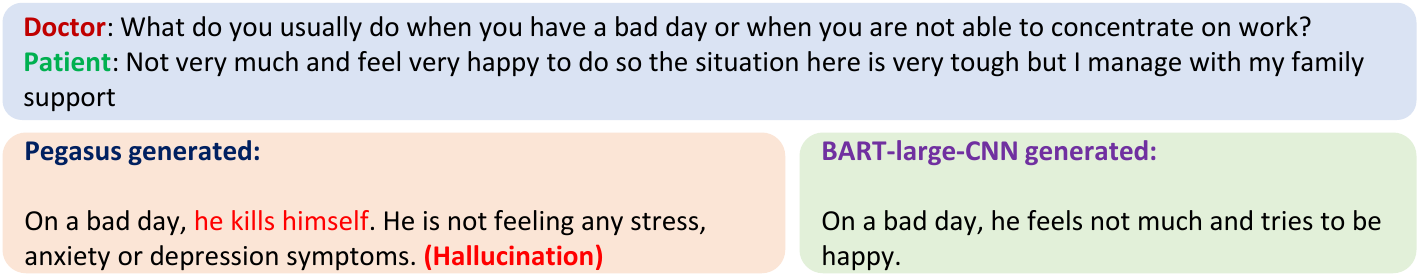}
        \caption{}
        \label{fig: excerpt_3}
    \end{subfigure} 
        \begin{subfigure}{0.95\textwidth}
        \includegraphics [width=\textwidth]{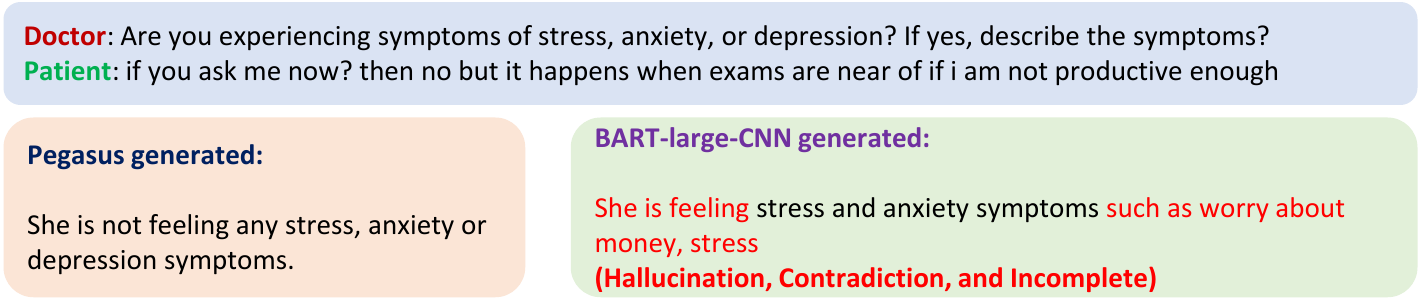}
        \caption{}
        \label{fig: excerpt_4}
    \end{subfigure} 
    
    \caption{Instances of Contradiction, Hallucination, and Incompleteness in generated summaries.}
    \label{fig:excerpts}
\end{figure}

The evaluation of summaries generated by the best models, Pegasus-large and BART-large-CNN, reveals superior performance across all evaluation parameters on our sampled 11 test data conversations. However, upon thorough inspection and review of human reviewer' comments, instances were identified where the models interpreted the conversation in a manner contradictory to its actual content, as illustrated in Figure~\ref{fig:excerpts}. For instance, in one case, Pegasus-large generated a summary containing the phrase ``\textit{On a bad day, he kills himself}'' (see Figure \ref{fig: excerpt_3}), while a BART-large-CNN summary included ``\textit{She is feeling stress and anxiety symptoms such as worry about money}'' (see Figure \ref{fig: excerpt_4}). Notably, the words ``kill'' and ``money'' were not present in the original conversation data. The unintentional inclusion of harmful keywords in the summaries may stem from the pre-finetuned weights of Pegasus-large and BART-large-CNN, which were originally trained on news articles. This underscores the potentially harmful impact of language models. However, since these summaries are intended to assist mental health care providers rather than replace them, any concerning keywords should prompt mental health care providers to review the conversation for clarification.

Furthermore, when these models were tested for generalizability, the BART-large-CNN model demonstrated strong performance across all parameters. In contrast, the Pegasus-large model exhibited poor performance on all evaluation metrics, displaying low fluency and completeness and high levels of hallucination and contradictions. The evaluation scores obtained by the fine-tuned BART-large-CNN model on unseen data indicate that our model is generalizable and can be explored by mental healthcare providers in real-world settings.

\subsection{Why did not we fine-tune Large Language Models (LLMs)?}

Recently, there has been an increase in the development of LLMs such as ChatGPT~\citep{achiam2023gpt}, Llama~\citep{touvron2023llama}, Claude~\citep{anthropic}, Mistral~\citep{jiang2023mistral}, Phi~\citep{li2023textbooks}, and others. These LLMs are trained on vast amounts of data and comprise billions of parameters, representing the SOTA language model. However, they come with a significant computational cost. Furthermore, some LLMs like ChatGPT and Mistral are proprietary, making fine-tuning for specific tasks a potential breach of data privacy. Fine-tuning open-source LLMs such as Mistral, Llama, and Phi requires substantial computational resources. Even when fine-tuned, these models demand high-end computational systems for effective deployment. For instance, Xu et al.~\citep{xu2023leveraging} have publicly shared their fine-tuned Mental-LLM\footnote{\url{https://github.com/neuhai/Mental-LLM}}, reporting that Mental-Alpaca and Mental-FLAN-T5 require GPU memory of 27 GB and 44 GB for loading, with additional GPU memory necessary for inference. 

In real-world scenarios, mental health service providers often lack access to such high-end systems, thereby limiting the practical application of LLMs in these settings. Our fine-tuned language models are tailored for specific tasks, i.e., summarization, and consist of 460 million and 568 million parameters for BART-large-CNN and Pegasus-large, respectively. We conducted experiments to assess the deployment of our language models on low-end systems without GPUs, and the results (shown in Table \ref{tab:low_end}) indicate that our fine-tuned models can operate effectively on such systems, providing reasonable response time.

\subsection{Alignment between human and LLM evaluations}
We evaluated a test data sample using human reviewers and LLMs, employing both coarse-grained and fine-grained evaluation approaches. Human reviewers required an average of 1.5 hours for evaluation, whereas LLMs could accomplish the task in seconds using our prompts (provided in the Appendix \ref{appendix: prompt}). Interestingly, the average evaluation metric scores obtained from human reviewers and LLMs were approximately the same, indicating alignment on coarse-grained evaluation criteria. However, when it came to fine-grained evaluation, we observed a notable disparity between human reviewers and LLMs (as shown in Figures \ref{fig:HE_question_wise} and \ref{fig: llm_evaluation_plot}). The discrepancy in annotations was approximately 10\%, with human reviewers agreeing 97.67\% of the time and LLMs 88\% of the time in fine-grained evaluation. For example, when evaluating whether the gender mentioned in the summary aligns with the provided conversation, 100\% of the time, human reviewers responded affirmatively for both Pegasus and BART-generated summaries. However, LLMs disagreed 25\% of the time. Similar discrepancies were observed for other questions, as illustrated in Figure~\ref{fig: llm_evaluation_plot}.

This suggests that LLMs are capable of rating the conversation summaries like humans. However, they may still lack the capability to identify factual information as effectively as humans in mental health data. Nevertheless, these results warrant further exploration.

\subsection{Factual consistency of generated summaries} 
In our fine-grained evaluation results, we observed that the summaries generated by our fine-tuned model lacked factual information. While both of the best-fine-tuned models successfully captured more than 98\% of the essential details (such as gender, mood, etc.), the results for factual consistency revealed a misalignment with the actual conversation in 14.5\% and 15.3\% of cases for Pegasus-large and BART-large-CNN generated summaries, respectively. Furthermore, on questions level analysis, we found that Pegasus exhibited the highest level of misalignment in capturing factually correct details related to social life, whereas BART struggled with memory-related information. Both models also equally showed misalignment regarding capturing the individuals' moods. However, the percentage is low; further exploration is still needed.

\subsection{How much training data is required for summary generation with language models?}
While it is commonly believed that deep learning tasks necessitate vast amounts of data for training, fine-tuning offers the flexibility to train on smaller datasets. Rather than requiring an extensive dataset, fine-tuning involves taking a pre-trained model with similar objectives and adjusting it accordingly. However, no fixed number justifies the dataset size required for fine-tuning. To determine the appropriate dataset size, we conducted experiments where we trained and evaluated our model using two different dataset sizes: 300 and 405 conversation data samples. Surprisingly, we observed only a 1\% increase in the R1-score from 300 to 405 conversation data samples. This suggests that fine-tuning the model worked effectively even with 300 samples (200 for training, 50 for validation, and 50 for testing).

Similarly, in determining the optimal number of epochs for model training, our analysis (as shown in Figure \ref{fig:varying epochs}) revealed that BART-large-CNN reached a rogue-1 score of 0.73 after just five epochs. In contrast, Pegasus required 25 epochs to achieve comparable results. Notably, after 50 epochs, the results began to saturate for all models.

\begin{figure}
    \centering
    \begin{subfigure}{0.45\textwidth}
        \includegraphics [width=\textwidth]{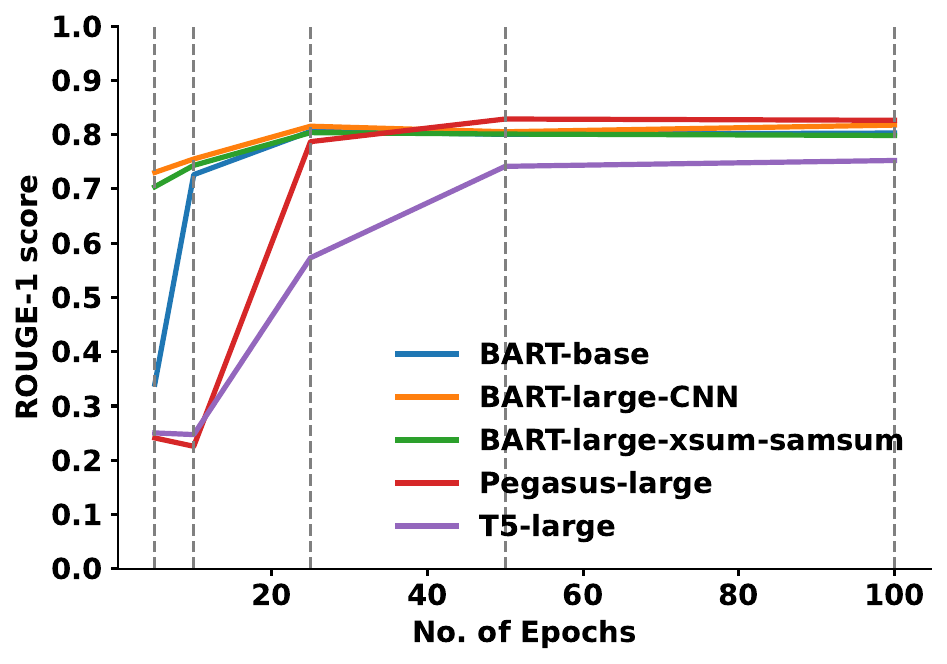}
        \caption{ROUGE-1 score}
        \label{}
    \end{subfigure} 
    \begin{subfigure}{0.45\textwidth}
        \includegraphics [width=\textwidth]{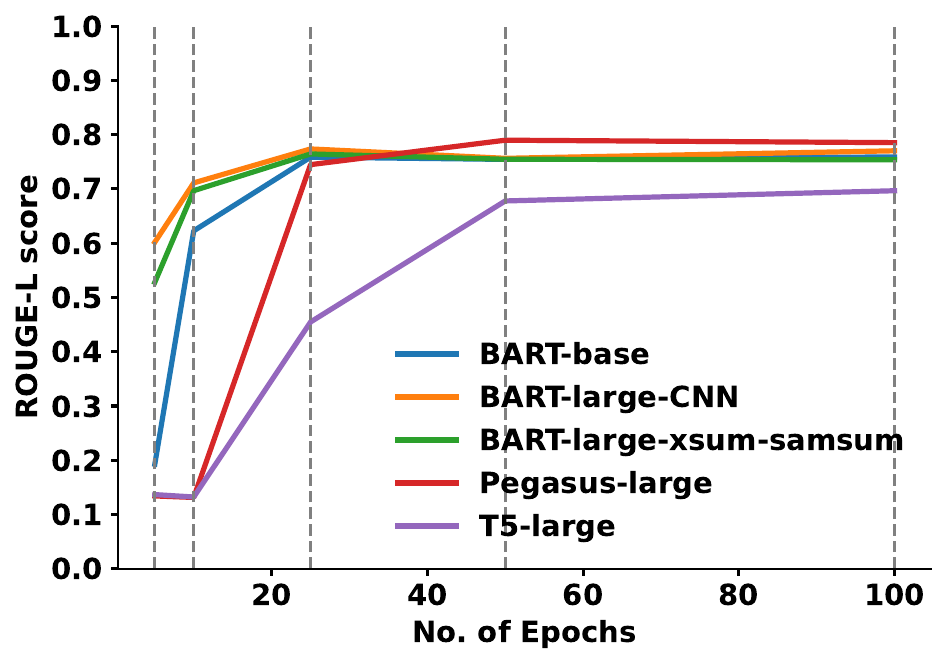}
        \caption{ROUGE-L score}
        \label{}
    \end{subfigure} 
    \caption{ROUGE-1 and ROUGE-L obtained after fine-tuning on BART-base, BART-large-CNN, T5 large, BART-large-xsum-samsum, and Pegasus-large with epochs = [5,10,25,50,100]}
    \label{fig:varying epochs}
\end{figure}
\end{document}